\documentclass[runningheads]{llncs}
\usepackage{graphicx}

\usepackage{tikz}
\usepackage{comment}
\usepackage{amsmath,amssymb} 
\usepackage{color}

\usepackage[accsupp]{axessibility}  

\usepackage[width=122mm,left=12mm,paperwidth=146mm,height=193mm,top=12mm,paperheight=217mm]{geometry}

\usepackage{booktabs}
\usepackage{algorithm}
\usepackage{algpseudocode}
\usepackage{mathtools}
\usepackage[normalem]{ulem}
\usepackage{makecell}
\usepackage{vcell}
\usepackage{multirow}
\usepackage{hhline}
\usepackage{caption}

\makeatletter
\def\blfootnote{\xdef\@thefnmark{}\@footnotetext}
\makeatother

\begin{document}
\pagestyle{headings}
\mainmatter

\title{Social-Implicit: Rethinking Trajectory Prediction Evaluation and The Effectiveness of Implicit Maximum Likelihood Estimation} 
\titlerunning{Social-Implicit: Rethinking Trajectory
Prediction Evaluation}
%
\author{Abduallah Mohamed\inst{1}\and
Deyao Zhu\inst{2} \and
Warren Vu\inst{1} \\
Mohamed Elhoseiny\inst{2,*} 
Christian Claudel\inst{1,*}
}
\authorrunning{A. Mohamed et al.}
%
\institute{The University of Texas at Austin\\
\email{\{abduallah.mohamed, warren.vu, christian.claudel\}@utexas.edu}\and
KAUST\\
\email{\{deyao.zhu, mohamed.elhoseiny\}@kaust.edu.sa} \and 
* Equal advising}
\maketitle

\begin{abstract}
Best-of-N (BoN) Average Displacement Error (ADE)/ Final Displacement Error (FDE) is the most used metric for evaluating trajectory prediction models. Yet, the BoN does not quantify the whole generated samples, resulting in an incomplete view of the model’s prediction quality and performance. We propose a new metric, Average Mahalanobis Distance (AMD) to tackle this issue. AMD is a metric that quantifies how close the whole generated samples are to the ground truth. We also introduce the Average Maximum Eigenvalue (AMV) metric that quantifies the overall spread of the predictions. Our metrics are validated empirically by showing that the ADE/FDE is not sensitive to distribution shifts, giving a biased sense of accuracy, unlike the AMD/AMV metrics. We introduce the usage of Implicit Maximum Likelihood Estimation (IMLE) as a replacement for traditional generative models to train our model, Social-Implicit. IMLE training mechanism aligns with AMD/AMV objective of predicting trajectories that are close to the ground truth with a tight spread. Social-Implicit is a memory efficient deep model with only 5.8K parameters that runs in real time of about 580Hz and achieves competitive results.
\end{abstract}

\section{Introduction}
\label{sec:intro}
\begin{figure}[!tbp]
\begin{center}
\includegraphics[width=0.9\textwidth]{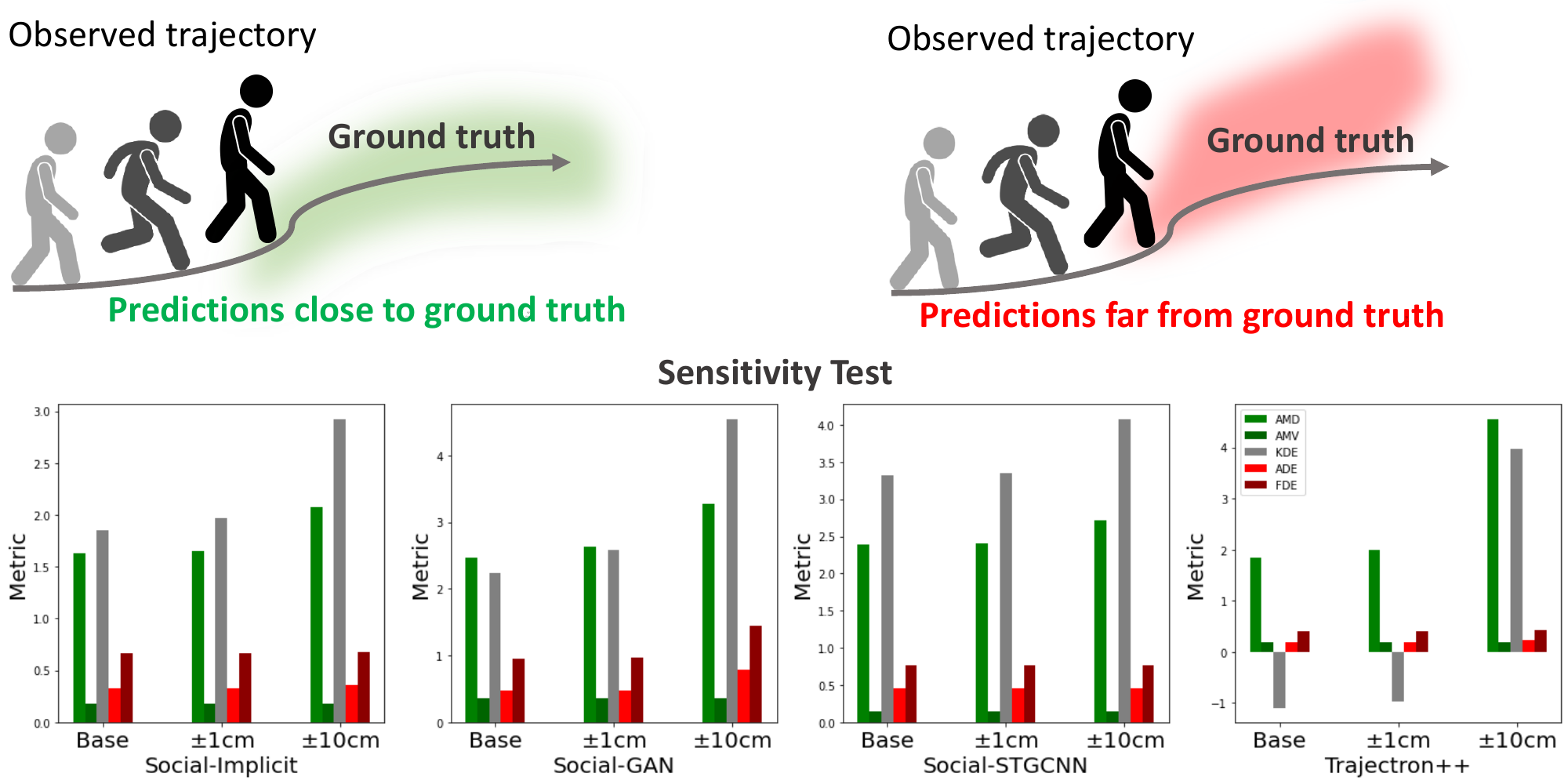}
\caption{The current BoN ADE/FDE metrics are not sensitive to the predicted distribution. The BoN ADE/FDE only focuses on the closest sample to the ground truth. We can see for both the green and red predictions, the BoN ADE/FDE stays the same. On the other hand, the proposed AMD/AMV metrics changes based on how close the whole predicted distribution and it is spread with respect to the ground truth. This makes the AMD/AMV a better metric to evaluate the predictions. }
\label{gr:social_implicit_teaser}
\end{center}
\end{figure}
\blfootnote{Code: \url{https://github.com/abduallahmohamed/Social-Implicit/}}
\blfootnote{Demo: \url{https://www.abduallahmohamed.com/social-implicit-amdamv-adefde-demo}}
Trajectory prediction is an essential component for multiple applications, such as autonomous driving~\cite{deo2018convolutional,li2019grip,cui2019multimodal,wu2020motionnet,rudenko2020human}, augmented reality~\cite{limmer2016robust,westphal2017challenges}, and robotics~\cite{liu2017human,butepage2018anticipating}. Typically, solving this problem requires a generative model to predict the future agent’s trajectories. Though there are plenty of deep models and design architectures that tackle this problem, the evaluation method used is being questioned. Typically, two metrics are used to evaluate the trajectory predictions models. The first one is Average Displacement Error (ADE)~\cite{pellegrini2009you} which is the average $L_2$ distance between the predicted and ground truth trajectories. Lower ADE values means that the overall predicted trajectory is close to the ground truth. The other metric is the Final Displacement Error (FDE)~\cite{alahi2016social}, which is an $L_2$ distance between the two final predicted and ground truth locations. In other terms, it describes if the predicted agent reaches its last goal or not. Also, the lower the FDE is, the better the model in not accumulating errors during the predictions. This issue of accumulating errors, resulting in a higher FDE was noticed in prior works that used recurrent based architectures. Prior works introduced the idea of a full CNN based architecture~\cite{mohamed2020social} to solve this error accumulation behavior.

Yet, this ADE/FDE metric remains unsuitable for generative models. Generative models predict multiple samples of future trajectories, implicitly forming a predicted distribution. This generative behavior is suitable for the problem, as the motion of an agent or pedestrian can be a multi-modal with possible future trajectories. In order to use the ADE/FDE in generative settings, the works of~\cite{alahi2016social,gupta2018social} introduced the concept of Best-of-N (BoN). BoN technique chooses from N samples the closest sample to the ground truth and calculates the ADE/FDE metric on it. This has a main issue of ignoring the set of generated samples. A model might generate an outlier sample that is luckily close enough to the ground truth, while the other samples are way off from the ground truth. This approach also fails in real-life applications, as there is a lack in the assessment of the predictions. Some important components, such as motion planning and collision avoidance, need a complete view of the predictions.
Another issue we noticed that the recent models~\cite{salzmann2020trajectron++,zhao2021you,mangalam2021goals,liu2021social} which are state-of-the-art based on the ADE/FDE metric only differ by 1cm ADE and few centimeters FDE on the ETH~\cite{pellegrini2009you} and UCY~\cite{lerner2007crowds} datasets, one of the most commonly used datasets in this area. The 1cm difference between a previous SOTA model and the next one is so subtle and tiny that it can be an annotation error or an outlier sampling. Thus, there is a need for a new metric that can evaluate the whole predicted samples and have a sense of where the whole generated distribution is regarding the ground truth. Also, there is a need to quantify the uncertainty of the generated samples giving a view regarding the confidence of the model, something that is needed in real-life applications. For this, we introduce the usage of Mahalanobis Distance~\cite{mahalanobis1936generalized} as a metric in this domain. We introduce two metrics, the Average Mahalanobis Distance (AMD) which evaluates how close a generated distribution is with respect to the ground truth, and the Average Maximum Eigenvalue (AMV) that evaluates the confidence of the predictions. The AMD quantifies how close a ground truth point is to a predicted distribution in a sense of standard deviation units. Also, AMD connects with $\chi^2$ distribution, helping us to determine the confidence of our predictions when the generated distribution degree of freedom is known. The AMV depends on the maximum magnitude of the eigenvalues of the covariance matrix of the predicted distribution. It quantifies the spread of the prediction. Thus, we can tell if a model is more confident than another model by using it. So, our goal is to achieve a model that generates a distribution which is close to the ground truth and has a small spread of samples around the ground truth. This aim leads us to rethink the nature of generative models used in training motion prediction models. We can classify the used generative techniques into parametric and non-parametric ones. Parametric ones use Maximum Likelihood Estimation (MLE) to model the predicted trajectories as Gaussian or Gaussian Mixture Models (GMM). Generative Adversarial Networks (GANs)~\cite{goodfellow2014generative} is an examples of non-parametric distributions. These approaches learn the distribution of the observed trajectories in order to generate the future ones. Yet, the primary goal of trajectory prediction models is the generated samples themselves. The MLE needs plenty of samples to converge, something we do not have in practice. While the GANs rely on the design of the discriminator and VAEs need to optimize the Evidence Lower Bound (ELBO). So, we needed a generative approach that only focuses on the generated samples and does not come with extra hassles. In this work, we show that Implicit Maximum Likelihood Estimation (IMLE) technique is an effective alternative to these approaches. IMLE focuses directly on the predicted trajectories, simplifying the optimization function. By using IMLE to train our introduced model Social-Implicit, the predicted trajectories improve in terms of quality and accuracy in comparison with prior works. Social-Implicit is a memory efficient deep model with only 5.8K parameters almost 55x less than the closest SOTA and runs in real-time almost 8.5x faster than the closest SOTA. This work is organized as follows: We start by literature review of recent relative works. Then we formulate the motion prediction problem, followed by an introduction and discussion for both new metrics AMD and AMV. Then we introduce the Trajectory Conditioned IMLE mechanism used in training our model Social-Implicit. We follow this by explaining the architecture of Social-Implicit. Lastly, we analyze the results of the new metrics on our model and recent SOTA ones accompanied with a sensitivity analysis.
\section{Literature Review}
\label{sec:literature}
\paragraph{Trajectory Forecasting Models}
Recent works have proposed various models to forecast future trajectories. Based on their output formats, they can be roughly grouped into two categories. Explicitly modeling the future as a parametric distribution, or implicitly modeling the future as a non-parametric distribution.
In the first category, methods model the future explicitly as continuous or discrete distribution~\cite{alahi2016social,mohamed2020social,rhinehart2019precog,chai2019multipath,salzmann2020trajectron++,tang2019multiple,cui2019multimodal,liang2020learning,zhu2021motion,lee2017desire,rhinehart2018r2p2,yuan2021agentformer,zhao2021you}. 
For example, S-LSTM~\cite{alahi2016social} and S-STGCNN~\cite{mohamed2020social} use Gaussian distribution to model the future trajectory that are trained by Maximum Likelihood Estimation (MLE). Gaussian distribution is single-mode and cannot catch the multi-modality of the future. To tackle this issue, PRECOG~\cite{rhinehart2019precog}, Trajectron++~\cite{salzmann2020trajectron++}, ExpertTraj~\cite{zhao2021you}, and AgentFormer~\cite{yuan2021agentformer} learn a latent behavior distribution, which can be either discrete~\cite{salzmann2020trajectron++,zhao2021you}
or continuous~\cite{rhinehart2019precog,yuan2021agentformer}, to represent the agent multi-modality intent. In these works, the predicted Gaussian distribution is generated conditioned on the sampled latent intent. This type of method is usually based on Conditional VAE~\cite{sohn2015learning}.
Besides continuous distributions like Gaussian methods like MTP~\cite{cui2019multimodal} and LaneGCN~\cite{liang2020learning} use a discrete distribution to represent the future. These methods predict a fixed number of deterministic trajectories as the future candidates and use a categorical distribution to model their possibilities. 
In the second category, some methods model the future distribution in an implicit way.
For example, S-GAN~\cite{gupta2018social}, SoPhie~\cite{sadeghian2019sophie}, S-BiGAT~\cite{kosaraju2019social} and DiversityGAN~\cite{huang2020diversitygan} follows a Conditional GAN~\cite{gauthier2014conditional} architecture. 
Instead of generating a distribution as the model output they predict a deterministic trajectory that is conditioned on a random sampled noise and is trained by an adversarial loss mechanism. Our proposed method Social-Implicit models the future distribution implicitly by training it using IMLE~\cite{li2018implicit} avoiding additional hassles like the discriminator in a GAN training mechanism.
\paragraph{Trajectory Forecasting Metrics}
Most of the trajectory forecasting methods are evaluated by the metric Average Displacement Error (ADE)~\cite{pellegrini2009you} or Final Displacement Error (FDE)~\cite{alahi2016social}. 
These two metrics are based on the $L_2$ distance of the whole temporal horizon (ADE) or the last time step (FDE) between the prediction and the ground truth trajectory. When the model generates a distribution as the output, the Best-of-N trick~\cite{gupta2018social} is applied to evaluate the best trajectory only from N sampled predictions. The mean ADE/FDE can be also used to evaluate the predictions, it is mostly suitable in single modality predictions and when the predictions are close to a Gaussian distribution. In multi-modality, when for example the predictions are contradictory to each other (turning left, turning right) the mean ADE/FDE will fail because it is deterministic in nature. Another way to evaluate the distribution quality is Kernel Density Estimate (KDE), first used in \cite{ivanovic2019trajectron}. KDE fits a kernel-based distribution from the prediction samples and estimates the negative log-likelihood of the ground truth as the evaluated score. Quehl et al.~\cite{quehl2017good} propose a synthesized metric that is a weighted sum of different similarity metrics to alleviate the metric bias. But their metric is only suitable for deterministic models. We propose two new metrics Average Mahalanobis Distance (AMD) and Average Maximum Eigenvalue (AMV) that are a better alternative for the BoN ADE/FDE in evaluating the predictions.

\section{The Average Mahalanobis Distance (AMD) metric}
We define the problem of trajectory prediction as follows: Given an observed trajectories of $N$ agents across a sequence of observed time steps $T_o$, the goal is to predict the next $T_p$ prediction time steps. The observed trajectories contain $P$ points, where each point indicates the spatial location of an agent. In the pedestrian trajectory prediction problems the $P$ is a 2D Cartesian locations (x,y). We denote the set of observation to be $d_o=\{p_t|\;t \in T_o \}$ and the set of predictions to be $d_p= \{p_t|\;t \in T_p \}$.

\begin{figure}[!tbp]
  \centering
  \begin{minipage}[b]{0.49\textwidth}
    \includegraphics[width=\textwidth]{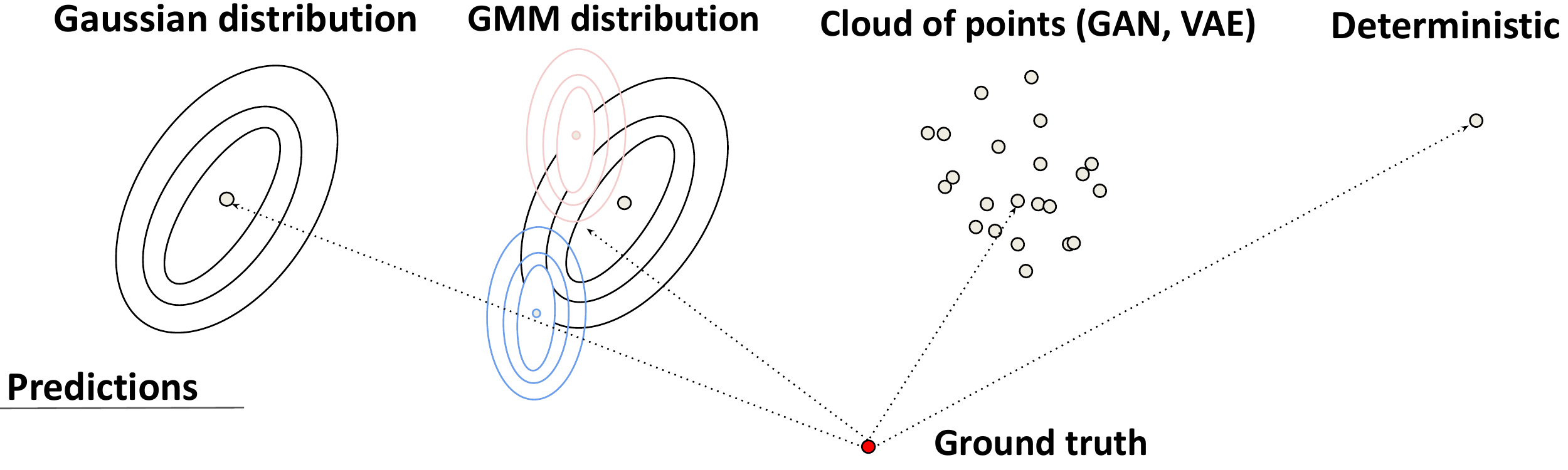}
   \caption{Different motion prediction models output. The Gaussian and GMM are examples of parametric models. The GAN and VAE are examples of non-parametric models. The last category is a deterministic model output. A unified metric is needed to evaluate all of these models.}
\label{gr:social_implicit_different_modes}
  \end{minipage}
  \hfill
  \begin{minipage}[b]{0.49\textwidth}
    \includegraphics[width=\textwidth]{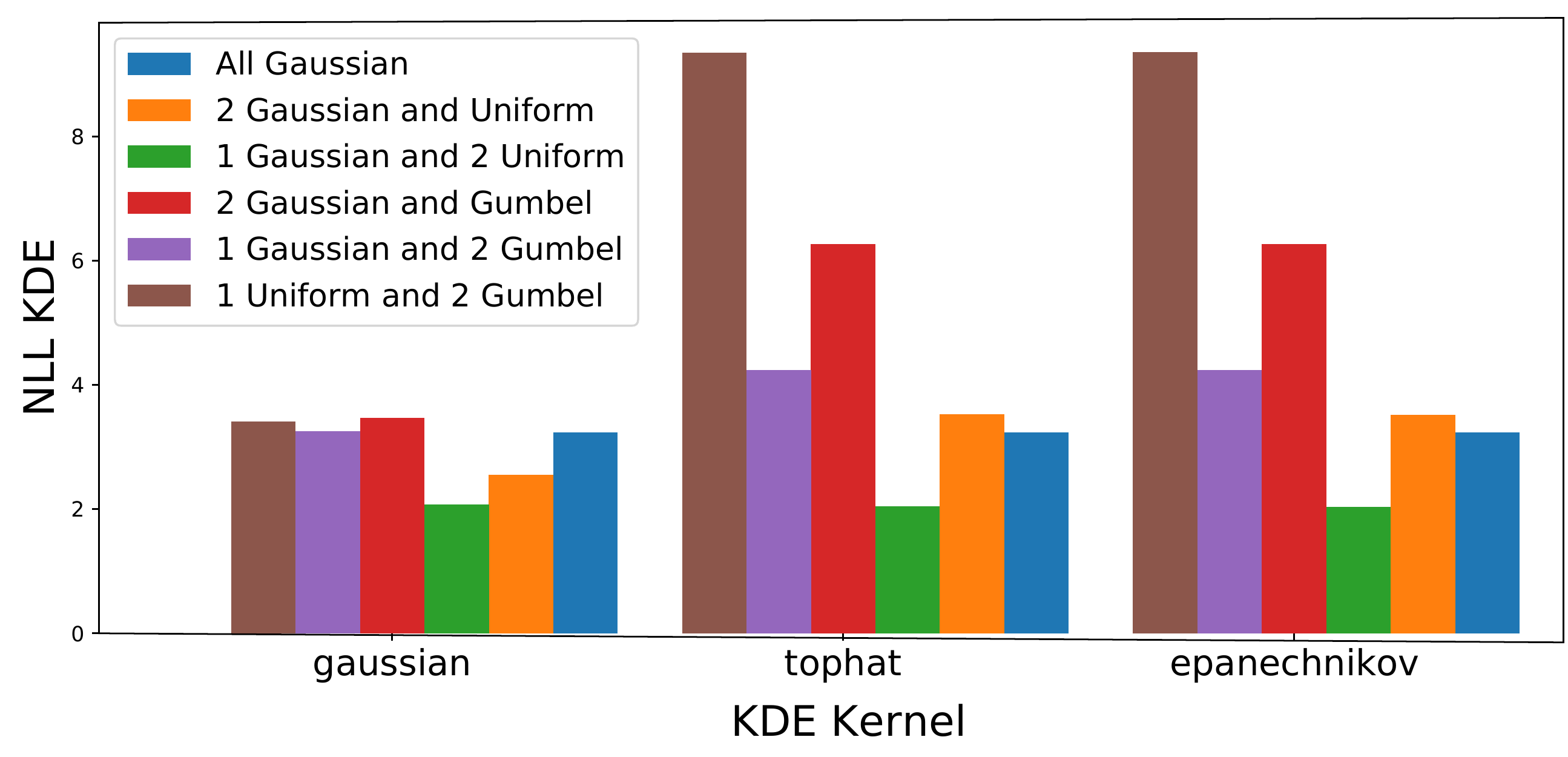}
       \caption{Different mixture models NLL KDE vs the choice of the KDE Kernel. The lower the better. We also notice that the results vary depending on the kernel.}
    \label{gr:social_implicit_kde_issue}
  \end{minipage}
\end{figure}

\begin{figure}[!tbp]
  \centering
  \begin{minipage}[b]{0.49\textwidth}
    \includegraphics[width=\textwidth]{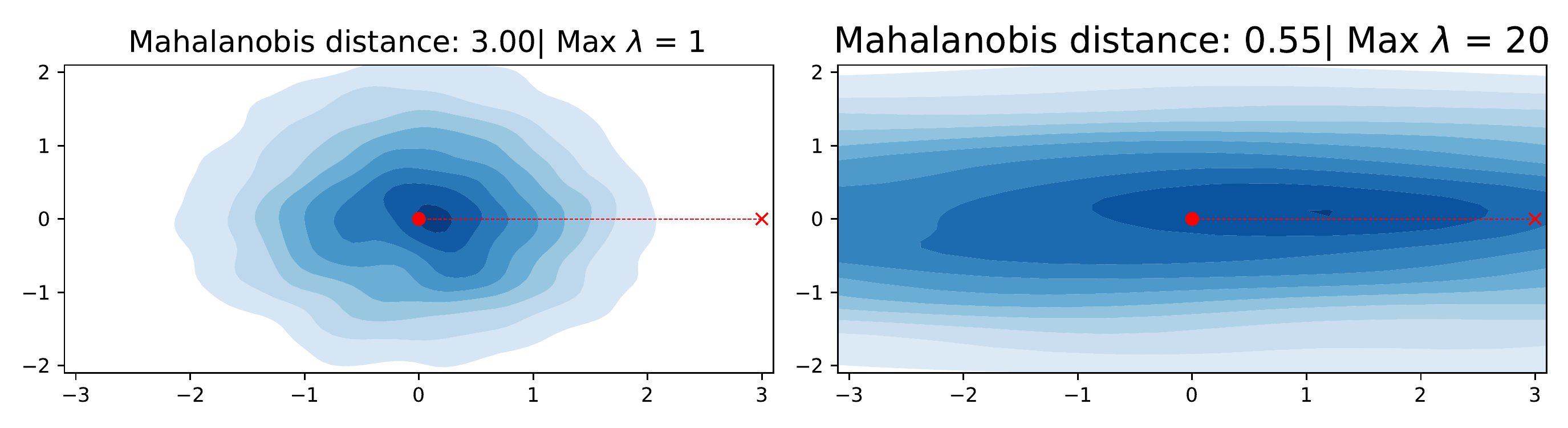}
       \caption{The Mahalanobis distance is being measured for a test point marked by x. Two Bi-Variate Gaussian distributions are shown, the one on the left has a lower variance than the one on the right. $\lambda$ stands for the maximum absolute eigenvalue of the distributions covariances.}
    \label{gr:social_implicit_md_issue}
  \end{minipage}
    \hfill
  \begin{minipage}[b]{0.49\textwidth}
    \includegraphics[width=\textwidth]{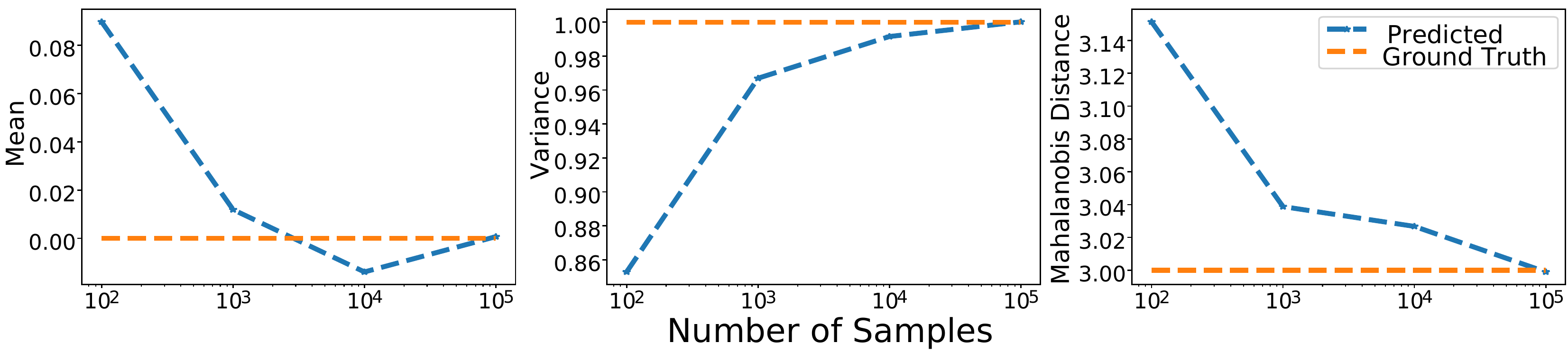}
       \caption{GMM fit exhibit an error and need a lot of samples to converge into the true mean, something might be a disadvantage to non-parametric models. At the 1000 samples mark, GMM starts to converge to the true mean and variance with stable MD.}
    \label{gr:social_implicit_gmm_example}
  \end{minipage}
\end{figure}

To highlight the issue in the current BoN ADE/FDE, we start with Fig.~\ref{gr:social_implicit_different_modes} that illustrates the different types prediction models outputs. For deterministic models, it is straightforward to compute the ADE/FDE metrics defined in Equation~\ref{eq:ADE_FDE_KDE}. But for generative models, the ADE/FDE is being computed by the BoN approach. The BoN works by sampling N (usually 20) samples, selecting the closest sample to the ground truth, then using this sample to calculate the ADE/FDE. We can criticize this BoN approach in multiple aspects. The major concern is that it does not quantify the whole generated samples and only focuses on the closest one. This might disadvantage a model with a density that is surrounding the ground truth against another model with a density that is completely off the ground truth, but has one sample that is close to the ground truth. We can see this illustrated in the teaser Fig.~\ref{gr:social_implicit_teaser}. We base the other concern that with this method of BoN, one can run the metric a couple of times, getting a result that is 1 cm better than another model. In some extreme cases, a lucky random run might have a very low BoN ADE/FDE. The work of~\cite{ivanovic2019trajectron} noticed this issue and introduced the usage of A Kernel Density Estimate (KDE) defined in Equation~\ref{eq:ADE_FDE_KDE}. The KDE is a kernel based tool that gets a non-parametric representation of the predictions’ probability density. Then, the negative log likelihood of the ground truth is calculated and reported in logarithmic units (nats). Yet, there is a mix of limitations and concerns with the KDE metric. 
The main concern is that the KDE metric is sensitive to the choice of a kernel under the settings of low number of samples, which is the case in real-life datasets. Fig.~\ref{gr:social_implicit_kde_issue} illustrates the different choices of the kernel used in the KDE versus a variety of mixtures of distributions. We notice that when a Gaussian kernel is being used; it does not differentiate between different samples and might favour a model with a full GMM output in comparison with other outputs. We also notice, we might get a mixed results whenever a different choice of kernel is being used, such as with tophat kernel versus a Gaussian kernel. The work of~\cite{ivanovic2019trajectron} was using KDE metrics with a Gaussian kernel. The other limitation of the KDE kernel is that it does not contain analytical properties that are easy to interpret. This limitation is because of the non-parametric nature of the KDE. Such properties of interest might be the probability moments and the confidence intervals.
\begin{equation}
\label{eq:ADE_FDE_KDE}
    \resizebox{0.93\linewidth}{!}{$
    \text{ADE} = \frac {1}{N \times T_{p}} \sum\limits_{n \in N} \sum\limits_{t \in T_{p}}\lVert \hat{p}^n_{\text{t}}-p^n_{\text{t}} \rVert_2 \text{,  }     \text{FDE} = \frac {1}{N} \sum\limits_{n \in N}\lVert \hat{p}^n_{ T_{p}}-p^n_{ T_{p}} \rVert_2 \text{,  } 
    \text{KDE} = \frac {-1}{N \times T_{p}}\sum\limits_{n \in N} \sum\limits_{t \in T_{p}} \log \text{KDE}(\hat{p}^n_{\text{t}},p^n_{\text{t}}) \quad \quad \quad \quad \quad \quad $}
\end{equation}
Where $p^n_{t}$ is ground truth location of agent $n \in N$ at predicted time step $t \in T_p$ and $\hat{p}^n_{t}$ is the predicted location.
The new metric needs to be parametric that allows further analysis and insensitive to the way it calculates the distance. Thus, we introduce the usage of Mahalanobis distance. Mahalanobis distance can measure how far a point from a distribution is, while correlating the distance with the variance of the predictions. It also has analytical properties that connect it with the Chi-square distribution, in which one can evaluate the confidence of the predictions. Lastly, it depends on Gaussian distribution, which allows further analysis of the predicted moments. The Mahalanobis distance (MD) is defined as:
$
    \label{eq:MD}
    M_D\left(\hat{\mu},\hat{\Sigma},p\right) = \sqrt{\left(p-\hat{\mu}\right)^T \hat{\Sigma}^{-1} \left(p-\hat{\mu}\right)}
$.
Where, $\hat{\mu}$ is the mean of the prediction, $\hat{\Sigma}$ is the variance of the predicted distribution and $p$ is the ground truth location.
Originally, Mahalanobis distance was not designed for a GMM distribution. Yet, the work of~\cite{tipping1999deriving} extended MD into a GMM by formulating it as:
\begin{equation}
    \label{eq:MD_MX}
    M_D\left(\hat{\mu}_{\text{GMM}},\hat{G},p\right) = \sqrt{\left(\hat{\mu}_{\text{GMM}} - p\right)^T \hat{G} \left(\hat{\mu}_{\text{GMM}} -p\right)}
\end{equation}
Where $\hat{G}$, the inverse covariances of each mixture components averaged and weighted probabilistically, is defined as: 
$
    \hat{G} = \frac{\sum^K_{k=1}\hat{\Sigma}^{-1}_k \hat{\pi}_k \int_{\hat{\mu}_{\text{GMM}}}^p p\left(x|k\right) dx }{\sum^K_{k=1}\hat{\pi}_k \int_{\hat{\mu}_{\text{GMM}}}^p p\left(x|k\right) dx}
$, where $K$ is the number of mixture components, $\hat{\pi}_k$ is the weight of the kth component and the mean of the GMM is defined as: $
    \hat{\mu}_\text{GMM}=\sum^K_{k=1} \hat{\pi}_k \hat{\mu}_k
$. The integral term in $\hat{G}$ is tractable, as noted in~\cite{tipping1999deriving}. We notice that if the GMM contains only one component, the $G$ will be the $\hat{\Sigma}^{-1}$, thus the Tipping's MD is a more generalized version of the original MD. Our approach is the following, whatever distribution or output produced by a model, we fit into a GMM. A question will be raised regarding the number of optimal mixture components $K$. This can be easily solved by using the Bayesian information criterion (BIC):
$
    \text{BIC} = m \ln{n} - 2 \ln{\hat{L}_{\text{GMM}}}
$, where m is the number of parameters of the GMM model, n is the number of observed data points and $\hat{L}_{\text{GMM}}$ is the likelihood function of the model. The lower the BIC is the better the fitted GMM model representing the data points. The best GMM is chosen automatically based on the BIC. Looking into the reason for sampling from a model that is already predicting the mean and variance of trajectories such as~\cite{alahi2016social,mohamed2020social}, that we want to be fair. Fitting a GMM will carry out a sort of error, thus we want this error to be incorporated into all modes of measurements to have a unified metric. Fig.~\ref{gr:social_implicit_gmm_example} show this error.

Because a deterministic model does not have a variance, we need a representation of the error in the model. We can train the deterministic model multiple times and fit the predictions to a GMM. Another proposal is to calculate the ensemble mean and variance and directly apply the MD distance without the GMM fit. The later approach might have an error that is equivalent to the GMM fit error, making the metric more fair. In the supplementary, we discuss both cases. We believe that evaluating a deterministic model versus a generative model is an open question that needs a further research and it is a limitation similar to the KDE~\cite{ivanovic2019trajectron} limitation. Now, we define the Average Mahalanobis Distance (AMD):
\begin{equation}
    \label{eq:AMD}
    \text{AMD} =
    \frac{1}{N \times T_p} \sum_{n \in N} \sum_{t \in T_p}  M_D\left(\hat{\mu}_{\text{GMM},t}^n,\hat{G}_t^n,p_t^n\right) 
\end{equation}

\section{The Average Maximum Eigenvalue (AMV) Metric}
A major concern of the AMD metric is that it is highly correlated with the variance of the distribution. A model might predict future trajectories, with a huge on-practical variance having the ground truth close to the mean. This will lead to a very low AMD in comparison with another model with a higher variance. Another example is a model that predicts a huge variance that is in meters covering all the predicted points. This also will lead to an optimal AMD value. To counter this false behavior, we need our models to have a low AMD accompanied with a low variance aka a more certain model. Also, in practical application, we need to quantify the overall uncertainty of the predictions to have a holistic view of the performance. Thus, we introduce the usage of the eigenvalues of the covariance matrix. The largest magnitude eigenvalue of the covariance matrix is an indicator of the spread of the covariance matrix. Fig.~\ref{gr:social_implicit_md_issue} illustrates two distributions, the one on the left has a smaller variance than the one on the right. We notice that the MD of a fixed point with respect to the left distribution is much higher when compared to the right distribution. Yet, the largest magnitude eigenvalue of the left distribution is way less than the right distribution, showing the spread of the predictions. So, to properly evaluate the models we need both of the AMD and a measurement of the spread. And as we discussed we can have a measurement of the spread directly from the prediction covariance matrix. Because of the framework we introduced in the AMD metric, we have a covariance matrix of the prediction. Something that was missing in the KDE metric. Now, we can introduce the AMV metric:
\begin{equation}
    \label{eq:AMV}
    \text{AMV} =
    \frac{1}{N \times T_p} \sum_{n \in N} \sum_{t \in T_p}
    \lambda^\downarrow_1(\hat{\Sigma}_{\text{GMM},t}^n)
\end{equation}
Where $\lambda^\downarrow_1$ is the eigenvalue with the largest magnitude from the matrix eigenvalues. The $\hat{\Sigma}_\text{GMM}$ is the covariance matrix of the predicted GMM distribution defined as:$
 \hat{\Sigma}_\text{GMM} = \sum_{k=k}^K \hat{\pi}_k \hat{\Sigma}_k + \sum_{k=1}^K \hat{\pi}_k(\hat{\mu}_k-\hat{\mu}_{\text{GMM}})(\hat{\mu}_k-\hat{\mu}_{\text{GMM}})^T
$. Thus, the AMV becomes a metric that evaluates the overall spread of the predicted trajectories. A model with low AMD will have a predicted distribution that is closer to the ground truth. And a model with low AMV will be more certain in their predictions. \textit{Thus, a model with both low AMD/AMV average is preferred when compared with another model with a higher AMD/AMV average.} For this we use the $\frac{AMD+AMV}{2}$ as an indicator of a good model. 

\section{Trajectory Conditional Implicit Maximum Likelihood Estimation (IMLE) Mechanism}
By induction from the goal of the AMD/AMV metric to have a model that generates samples that is close to the ground truth with low spread, we need a training mechanism that allows full control over the predicted samples as the main optimization goal. Typical training mechanism such as Maximum Likelihood Estimation (MLE) or its variants like maximizing evidence lower bound (ELBO) encourages the prediction samples to be close to some (ground truth) data sample. In this way, some data examples may be missed and lead to a mode dropping~\cite{li2018implicit}. Other methods, such as GANs, need to introduce additional modules like the discriminator and their training is usually unstable and need to be carefully tuned to reach a proper Nash's equilibrium. The work of~\cite{li2018implicit} introduced the concept of Implicit Maximum Likelihood Estimation (IMLE). IMLE encourages every target ground truth to be close to some predicted samples. Therefore, it leads to a predicted distribution that is better in covering the ground truth unlike MLE. IMLE trains a model via a simple mechanism: inject a noise into the model's input to predict multiple samples, select the one closest to the ground truth, and back propagate using this sample. Unlike other generative approaches, IMLE does not load the optimization objective with a specific training technique and keeps the training stable due to the simple distance-minimization-based optimization. Using IMLE as a training mechanism aligns with the AMD/AMV goals and focuses on the important product, the predicted output. Another point of view for IMLE is that it is a more advanced neural technique in comparison with estimation techniques such as Kalman filter where the process and measurement noises drive the model. We refer the reader to the original IMLE paper~\cite{li2018implicit} for a further discussion. The training mechanism is shown in Alg.\ref{alg:imle_proc}:
\begin{algorithm}[h]
\caption{Trajectory Conditional Implicit Maximum Likelihood Estimation (IMLE) algorithm}
\label{alg:imle_proc}
\begin{algorithmic}
\Require The dataset $D={(d_o^i,d_p^i)}_{i=1}^n$ and the model $\theta(.)$ with a sampling mechanism conditioned on the input
\Require Choose a proper loss function $\mathcal{L}(.)$ such as mean squared error or $L_1$
\State Initialize the model
\For{$e = 1$ \textbf{to} $\text{Epochs}$}
    \State Pick a random batch $(d_o, d_p)$ from $D$
    \State Draw i.i.d. samples $\tilde{d_p}^{1},\ldots,\tilde{d_p}^{m}$ from $\theta(d_o)$\
    \State $\sigma(i) \gets \arg \min_{i} \mathcal{L}\left( d_p-\tilde{d_p}^{i}\right)\;\forall i \in m$
    \State $\theta \gets \theta - \eta \nabla_{\theta}\sigma(i)$
\EndFor
\State \Return $\theta$
\end{algorithmic}
\end{algorithm}

\section{The Social-Implicit Model}
\begin{figure}[t]
\begin{center}
\includegraphics[width=0.9\linewidth]{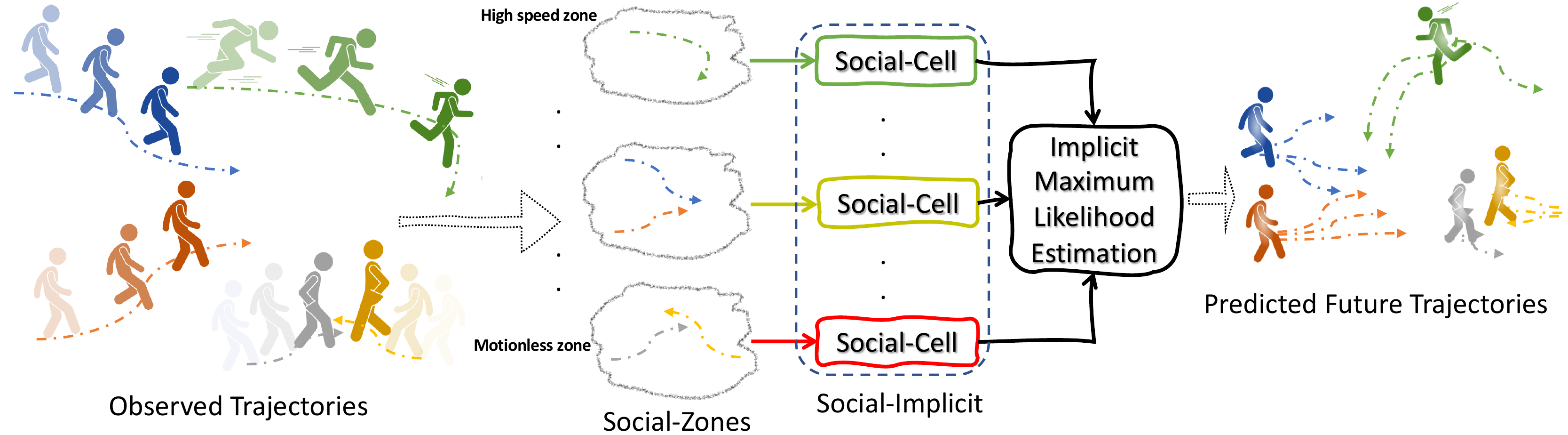}
   \caption{Social-Implicit model concept. The Social-Zones cluster the observed trajectories based on their maximum observed speed. Then each Social-Zone is processed by a Social-Cell. The model is trained using IMLE. }
   \end{center}
\label{gr:social_implicit_model}
\end{figure}
\begin{figure}[t]
\begin{center}
\includegraphics[width=0.7\linewidth]{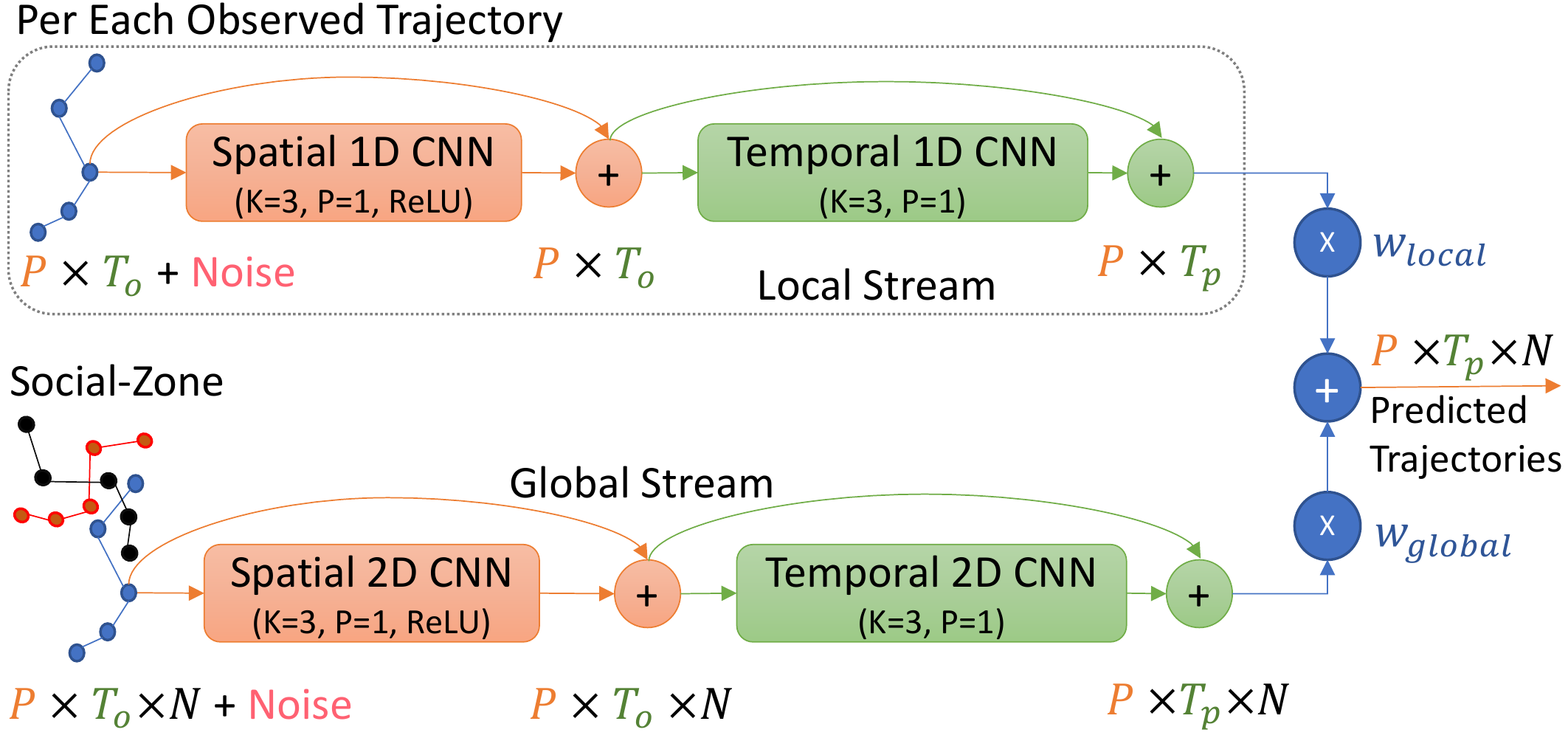}
   \caption{Social-Cell model. The local and global stream has only two CNNs. $P$ is the observed location, $T_o$ and $T_p$ is the observed and predicted time steps and $N$ is the count of agents. Where $K$ and $P$ of the CNNs is the kernel and padding size.}
\label{gr:social_implicit_social_cell}
   \end{center}
\end{figure}

In this section, we present the Social-Implicit model. The Social-Implicit is tiny in memory size with only 5.8K parameters with real run-time of 588Hz. The method comprises three concepts, Social-Zones, Social-Cell and Social-Loss.\\
\noindent\textbf{The Social-Zones: }The Social-Zones cluster the observed agents trajectories based on their maximum change of speed. The average pedestrian speed is ~1.2m/s~\cite{laplante2004continuing}. We noticed that we can cluster the motion of pedestrians into four groups. The first group is the motion-less group, where the pedestrian is waiting at the traffic light as an example. This group’s maximum speed change is between 0-0.01m/s. While the second group is pedestrians with minimal motion, aka someone who is shaking in place or a group of pedestrians greeting each other, typically this group's maximum change of speed is between 0.01-0.1m/s. The third group are pedestrians with an average walking speed, these pedestrians motion is between 0.1-1.2m/s. The last group is the running pedestrians, typically with a speed above the 1.2m/s average. When a deep model is trained on the stationary pedestrians alongside the faster ones, a bias towards the moving ones will exist in the predictions. This will force the model to predict the non-moving objects as moving ones. It is a sort of data imbalance, or in other terms, a zero(motionless)-inflated data issue. Hence, the concept of Social-Zones is needed to solve this issue. Empirically, we show that our model with the Social-Zones performs better than without it. The input to the Social-Zones is the observed trajectories and the output is clusters of pedestrians, each cluster is a graph of dimensions $P\times T_o \times N$.\\
\noindent\textbf{The Social-Cell: }The fundamental building unit of the Social-Implicit model is the Social-Cell. The Social-Cell is a 4 layers deep model that is simple and directly dealing with the spatio-temporal aspects of the observations. Fig.~\ref{gr:social_implicit_social_cell} illustrates the structure of the Social-Cell. We notice the cell has two components, one that deals with each individual agent at a local level and one that deals with the whole agents at a global level. We generate the final output of the cell by combining the local and global streams via self-learning weights. Both local and global streams are two consecutive residually connected CNN layers. The first CNN is a spatial CNN, which creates an embedding of the spatial information of the observed agents. The second layer is a temporal CNN that treats the time aspect of the observed trajectories. It treats the time as a feature channel, allowing us to predict the next $T_p$ time steps without using a recurrent network~\cite{mohamed2020social}. We found out that this simple architecture is as effective as much larger and complex models, resulting in a small memory size and real-run time capabilities. Each Social-Cell deals with a specific Social-Zone. The input is $P\times T_o \times N$ and the output is $P\times T_p \times N$. The operations are shown in Fig.~\ref{gr:social_implicit_social_cell}.
\textbf{The Social-Loss: }The loss function of Social-Implicit exhibits several parts. The first part is the direct optimization objective of the IMLE mechanism that we discussed before. The second part is a triplet loss. This triplet loss considers the anchor to be the closest sample $\tilde{d}^1_p$ to the ground truth. The positive example is the next closest example $\tilde{d}^2_p$ to the ground truth. The negative example $\tilde{d}^m_p$ is the farthest sample from the ground truth. This helps in grouping the samples closer to the ground truth, resulting in a tighter distribution around the real trajectory. The last part of the loss is a geometric loss function that treats the predicted locations as a polygon. First, it ensures that the intra-distance between the predicted location matches the intra-distance between the ground truth locations. Second, it makes sure that the angles between the predicted points are the same as the angles between the ground truth points. It ensures that the predicted scene geometrically looks like the ground truth. We defines these losses in Equation~\ref{eq:triplet_geometric_loss}. The social aspects of the scene can be addressed beyond what we introduced which is an open research area. 
\begin{gather}
\label{eq:triplet_geometric_loss}
\begin{aligned}
\mathcal{L}_{\text{triplet}} =\Vert \tilde{d}^1_p - \tilde{d}^2_p \Vert_1 - \Vert \tilde{d}^1_p - \tilde{d}^m_p \Vert_1 \quad  \quad  \quad  \quad \quad &\\
\mathcal{L}_{\text{G-distance}} =\frac{1}{\frac{T_p (T_p-1)}{2}}  \sum_{t \in T_p} \sum_{j \in T_p, j>t} \big\Vert \Vert p_t - p_j \Vert_2 - \Vert \tilde{p}_t - \tilde{p}_j \Vert_2 \big\Vert_1\\
\mathcal{L}_{\text{G-angle}} =\frac{1}{\frac{T_p (T_p-1)}{2}}  \sum_{t \in T_p} \sum_{j \in T_p, j>t} \big\Vert \angle (p_t,p_j)  - \angle (\tilde{p}_t,\tilde{p}_j)  \big\Vert_1
\end{aligned} 
\end{gather}

Thus we define the Social-Loss as: 
\begin{equation}
\mathcal{L} = \Vert d_p - \tilde{d}_p^1 \Vert_1 + \alpha_{1} \mathcal{L}_{\text{triplet}} +   \alpha_{1} \mathcal{L}_{\text{G-distance}} + \alpha_{3} \mathcal{L}_{\text{G-angle}}
\end{equation}
Where $\alpha_{1} = 0.0001 , \alpha_2= {0.0001 \text{ or }0.00001}, \alpha_3 = 0.0001$.
\section{Experiments and Analysis }
We analyze the behavior of the metrics on common pedestrian motion prediction models in terms of overall performance and sensitivity analysis. Then we analyze our model in terms of design components and performance. 
\subsection{Metrics Sensitivity Analysis \& Evaluation }
We show that the BoN ADE/FDE metric is not sensitive to the change or shift in the distribution, while the AMD and KDE can quantify such a change. Fig.~\ref{gr:social_implicit_teaser} illustrates this concept. We tested different models by shifting their predicted samples using different amounts, specifically $\pm1$cm and $\pm10$cm. In all of the models, the BoN ADE/FDE metric did not change at all or had a very tiny subtle change. Unlike the metrics that measure the whole distribution, like AMD and KDE the shift of the predicted distribution is reflected into the metric. This concludes that the ADE/FDE metric is not sensitive to the change of the whole distribution, even on a tremendous change of 10cm, which sometimes can define a new SOTA model over another. So, the BoN ADE/FDE metric is incapable of evaluating the whole predicted trajectories. Also, the AMV metric stayed the same, this was expected as only shifting the predictions does not change the variance. We notice that the KDE of Trajectron++ is -ve unlike other models because Trajectron++ output is a GMM distribution which is a bias in the KDE metric due to the kernel choice as we discussed earlier.

To evaluate the metrics quantitatively, we report the AMD/AMV, KDE and ADE/FDE metrics on different motion prediction models using the ETH/UCY datasets. We chose classic ones such as S-GAN~\cite{gupta2018social} and S-STGCNN~\cite{mohamed2020social}. We also chose more recent ones such Trajectron++~\cite{salzmann2020trajectron++} and ExpertTraj~\cite{zhao2021you}. From Tab.~\ref{tab:main_results} we notice that the last two models, which are considered SOTA, differ by a few centimetres on the ADE/FDE metric. Yet, when we evaluate both of them using the AMD/AMV metrics, we notice that Trajectron++ is performing much better than the ExpertTraj model. From the AMD/AMV metric, ExpertTraj generates a tight distribution that does not surround the ground truth, which results in a higher AMD unlike the Trajectron++. Though, both of ExpertTraj and Trajectron++ have very close ADE/FDE metrics, the quality of the whole predicted samples is completely different. Examining the results of our model, Social-Implicit, we see that it has the lowest AMD/AMV. By digesting the results, the ADE/FDE metric is not an indicative of the overall performance of the models which correlates with the aforementioned sensitivity analysis. We test our model and metrics on Stanford Drone Dataset (SDD)~\cite{robicquet2016learning}. We follow the setting of a SOTA model DAG-Net~\cite{monti2021dag}. Experimental results in Tab~\ref{tab:SDD_results} show that our model outperforms DAG-Net. This aligns with the results on the ETH/UCY datasets.

\begin{table}[!ht]
\centering
\tiny
\caption{For all metrics, the lower the better. The results are on the ETH/UCY datasets. M is a non reported model. NaN is failed computation. ExpertTraj ADE/FDE were taken from their paper. We notice sometimes, even if a model has a low ADE/FDE the AMD/AMV contradicts this by evaluating the overall generated samples.}
\label{tab:main_results}
\begin{tabular}{c||c|c|c|c|c||c||c|}
                               & ETH                        & Hotel                      & Univ                       & Zara1                       & Zara2                       & Mean                    & (AMD+AMV)/2  \\ 
\cline{2-8}
                               & \multicolumn{6}{c}{\makecell{ ADE/ FDE\\ AMD/ AMV\\KDE}    }                                                                                                                                                     &          \\ 
\hline
S-GAN~\cite{gupta2018social}    & \makecell{0.81/1.52 \\ 3.94/0.373 \\5.02}  & \makecell{0.72/1.61\\2.59/0.384\\3.45}  & \makecell{0.60/1.26\\2.37/0.440\\2.03}    & \makecell{0.34/0.69\\1.79/0.355\\0.68}    & \makecell{0.42/0.84\\1.66/0.254\\-0.03}  & \makecell{0.58/1.18\\2.47/0.361\\2.23}  & 1.42    \\ 
\hline
S-STGCNN~\cite{mohamed2020social} & \makecell{0.64/1.11\\3.73/0.094\\6.83}  &\makecell{ 0.49/0.85\\1.67/0.297\\1.56}  &\makecell{ 0.44/0.79\\3.31/0.060\\5.65}  &\makecell{ 0.34/0.53\\1.65/0.149\\1.40}   &\makecell{ 0.30/0.48\\1.57/0.103\\1.17}   &\makecell{ 0.44/0.75\\2.39/0.104\\3.32}  & 1.26  \\ 
\hline
Trajectron++~\cite{salzmann2020trajectron++}  & \makecell{0.39/0.83\\3.04/0.286\\1.34 }      &\makecell{ 0.12/0.21\\1.98/0.114\\-1.89}      &\makecell{ 0.20/0.44\\1.73/0.228\\-1.08}      &\makecell{ 0.15/0.33\\1.21/0.171\\-1.38}  &\makecell{ 0.11/0.25\\1.23/0.116\\-2.43}  &\makecell{ 0.19/0.41\\1.84/0.183\\\textbf{-1.09} }     & 1.01        \\ 
\hline
ExpertTraj~\cite{zhao2021you}    &\makecell{ 0.30/0.62\\61.77/0.034\\NaN}  &\makecell{ 0.09/0.15\\21.30/0.003\\NaN}   &\makecell{ 0.19/0.44\\M/M\\M}             &\makecell{ 0.15/0.31\\32.14/0.005\\NaN}    &\makecell{ 0.12/0.24\\M/M\\M}              &\makecell{ \textbf{0.17}/\textbf{0.35}\\38.4/0.004\\NaN}    & 19.20  \\ 
\midrule
\textbf{Social-Implicit(ours) }         &\makecell{ 0.66/1.44\\3.05/0.127\\5.08}  &\makecell{ 0.20/0.36\\0.58/0.410\\0.59}  &\makecell{ 0.31/0.60\\1.65/0.148\\1.67}  &\makecell{ 0.25/0.50\\1.72/0.078\\1.24}  &\makecell{ 0.22/0.43\\1.16/0.106\\0.69}   &\makecell{ 0.33/0.67\\\textbf{1.63}/\textbf{0.174}\\1.85} & \textbf{0.90}   \\
\hline
\end{tabular}
\end{table}

\begin{table}
\centering
\caption{For all metrics, the lower the better. The results are on the SDD dataset.}
\label{tab:SDD_results}
\begin{tabular}{c|ccccc|c|}
\toprule
        &   ADE &   FDE &   AMD &   AMV &   KDE & (AMD+AMV)/2 \\
\midrule
    STGAT~\cite{huang2019stgat} & 0.58 & 1.11 & - & - &- & -\\
    Social-Ways~\cite{amirian2019social} & 0.62 & 1.16& - & - &- & -\\    
    DAG-Net~\cite{monti2021dag} &  0.53 & 1.04 & 3.17 & 0.247 & \textbf{1.76}& 1.70  \\
    \midrule
    \textbf{Social-Implicit (ours)} & \textbf{0.47} & \textbf{0.89} & \textbf{2.83} & \textbf{0.077} & 3.89 & \textbf{1.45}\\
\bottomrule
\end{tabular}

\end{table}

\subsection{Ablation Study of Social-Implicit}
We conduct an ablation study of the Social-Implicit components. Specifically, the Social-Zones and the Social-Loss. Tab.~\ref{tab:social_implicit_ablation} illustrates the results. We noticed that the existence of the Social-Zones enhanced the AMD metric by almost 40\%. It also led to a good AMV value, which enhanced the overall AMD/AMV performance. We notice that the triplet loss alone with the Social-Zones leads improves AMD/AMV. The effect of geometric angle loss is more than the geometric distance loss in improving the AMD/AMV. While both work better together. 
\begin{table}
\centering
\tiny
\caption{Ablation study of Social-Implicit components.}
\label{tab:social_implicit_ablation}
\begin{tabular}{@{\hskip3pt}l@{\hskip3pt}l@{\hskip3pt}l@{\hskip3pt}l@{\hskip3pt}|l@{\hskip3pt}l@{\hskip3pt}l}
$\mathcal{L}_\text{triplet}$ & $\mathcal{L}_{\text{G-distance}}$ & $\mathcal{L}_{\text{G-angle}}$ & Zones & AMD/KDE/AMV       & AMD/AMV & ADE/FDE    \\ 
\hline
        &    &       &       & 2.06/2.29/0.110 & 1.09    & 0.32/0.66  \\
        &    &       & \checkmark      & 2.04/1.86/0.104 & 1.07    & 0.32/0.64  \\
\checkmark        &    &       & \checkmark      & 1.96/2.16/0.097 & 1.03    & 0.56/1.08  \\
\checkmark        & \checkmark   &       & \checkmark      & 2.13/2.32/0.092 & 1.11    & 0.32/0.62  \\
\checkmark        &    & \checkmark      & \checkmark      & 1.84/1.78/0.094 & 0.97    & 0.78/1.38  \\
\checkmark        & \checkmark   & \checkmark      &       & 2.29/2.76/0.090 & 1.16    & 0.35/0.71  \\
\checkmark        & \checkmark   & \checkmark      & \checkmark      & 1.63/1.85/0.174 & \textbf{0.90 }   & 0.33/0.67 \\
\hline
\end{tabular}

\end{table}

\subsection{Inference and Memory analysis}
Social-Implicit besides being the most accurate model when compared with other models on the AMD/AMV metrics, it is the smallest and the fastest in terms of parameters size and inference time. Table~\ref{tab:speed_and_parameters} shows these results. The closest SOTA is ExpertTraj which Social-Implicit is 55x smaller and 8.5x faster. 

\subsection{Social-Zones ablation} Tab.~\ref{tab:zones} shows the ablation of the number of zones. Different zones affects the model’s performance. Also, the model is sensitive to the zone's speeds. For example, when we changed the last zone from 1.2m/s to 0.6m/s the results changed. The 1.2m/s reflects human average walking speed, thus the 0.6m/s does not suit the data, hence it leads to poor performance in comparison with the 1.2m/s.
\begin{table}[!htb]
    \begin{minipage}{.48\textwidth}
    \tiny
    \caption{Parameters counts and mean inference speed reported, benchmarked on the GTX1080Ti.}
    \label{tab:speed_and_parameters}
      \centering
    \begin{tabular}{l|l|l}
                    & Parameters count & Speed (s) \\ \hline
    S-GAN~\cite{gupta2018social}      & 46.3K \textcolor{blue}{(7.98x)}                  & 0.0968 \textcolor{blue}{(56.9x)}              \\ \hline
    S-STGCNN~\cite{mohamed2020social}   & 7.6K  \textcolor{blue}{(1.3x)}                & 0.0020  \textcolor{blue}{(1.2x) }            \\ \hline
    Trajectron++~\cite{salzmann2020trajectron++}    & 128K  \textcolor{blue}{(22.1x)}                 & 0.6044  \textcolor{blue}{(355.5x)}             \\ \hline
    ExpertTraj~\cite{zhao2021you}      & 323.3k \textcolor{blue}{(55.74x)}                & 0.0144  \textcolor{blue}{(8.5x) }            \\ \hline
    \bottomrule
    \textbf{Social-Implicit (ours)} & \textbf{5.8K}              & \textbf{0.0017 }             \\ \hline
    \end{tabular}
    \end{minipage}%
    \hfill
    \begin{minipage}{.48\textwidth}
    \scriptsize
      \centering
        \caption{Number of zones and their speed effect on our model accuracy.}
        \label{tab:zones}
        \begin{tabular}{l|l|l|l|l}
        \#Zones              & ADE/FDE   & KDE  & AMD/AMV    & AVG  \\ 
        \hline
        1                  & 0.35/0.71 & 2.76 & 2.29/0.090 & 1.16           \\
        2                  & 0.36/0.73 & 3.09 & 2.32/0.088 & 1.20           \\
        3                  & 0.34/0.70 & 2.31 & 2.08/0.085 & 1.08           \\
        4@0.6m/s & 0.34/0.69 & 2.85 & 2.31/0.080  & 1.19           \\ 
        \hline
        4@1.2m/s& 0.33/0.67 & 1.85 & 1.63/0.174 & \textbf{0.90}\\
        \hline
        \end{tabular}
    \end{minipage} 
\end{table}
\subsection{Qualitative Results}
\begin{figure}[h]
\centering
\includegraphics[width=0.8\columnwidth]{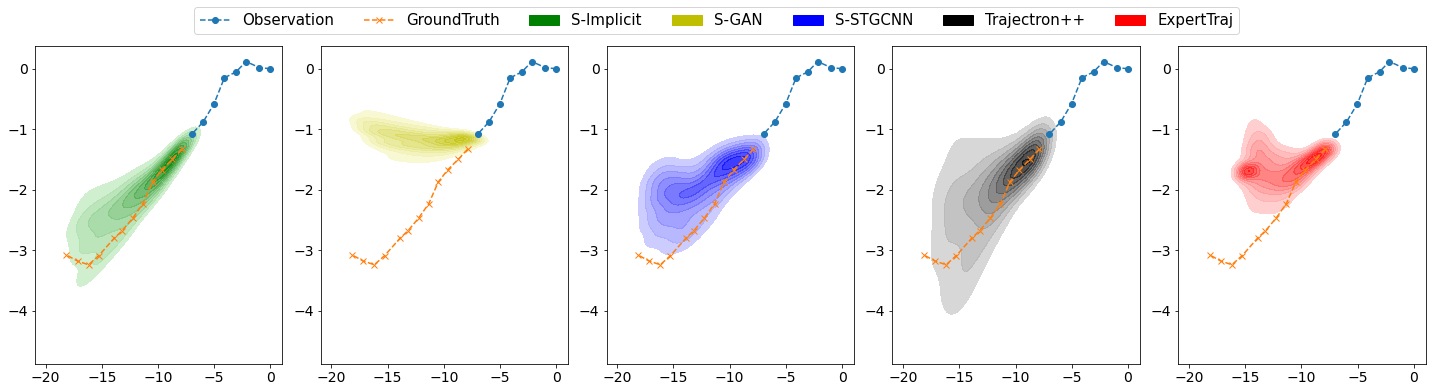}
\centering
\includegraphics[width=0.8\columnwidth]{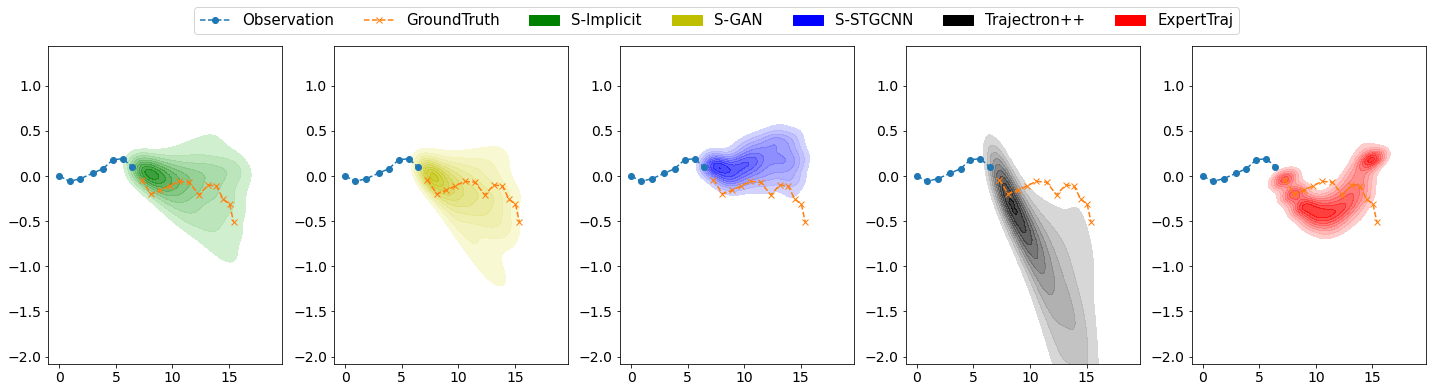}
\caption{Visualization of the predicted trajectories on the ETH/UCY datasets.}    
\label{fig:eth_visual}
\end{figure}

\begin{figure}[h]
\centering
\includegraphics[width=0.8\columnwidth]{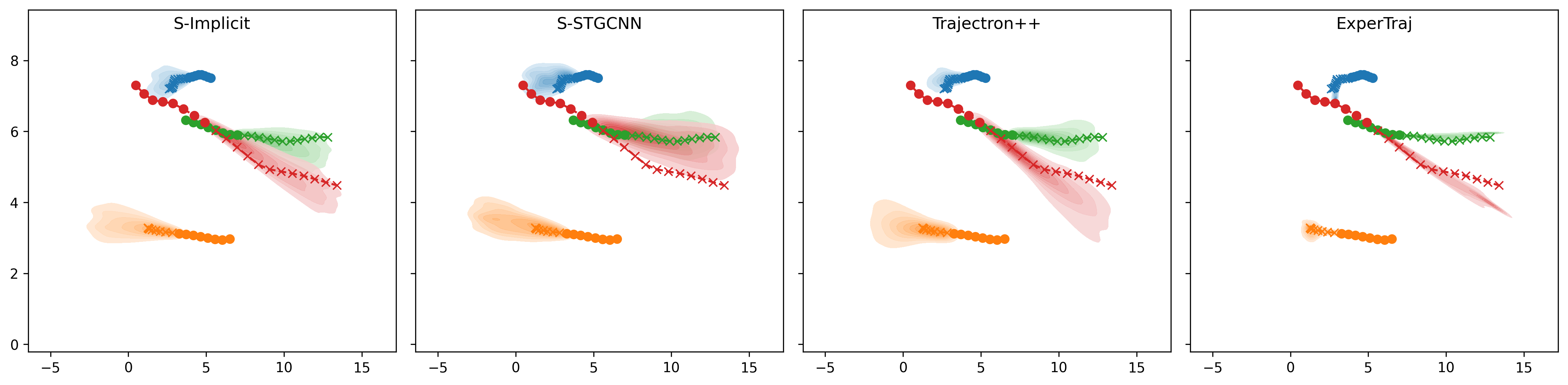}
\centering
\includegraphics[width=0.8\columnwidth]{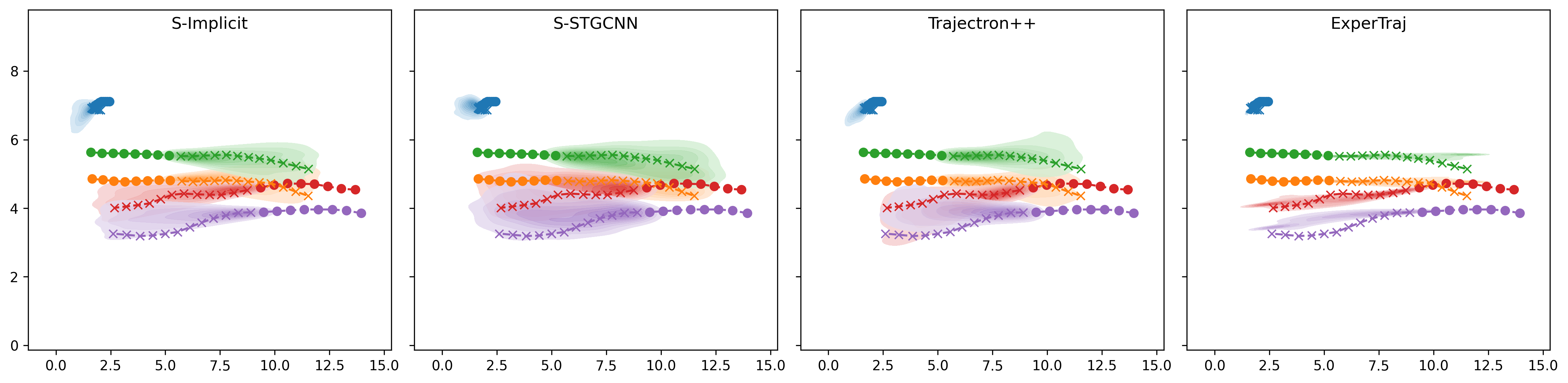}
\caption{Multi-pedestrian interaction cases on the ETH/UCY datasets.}    
\label{fig:interaction}
\end{figure}

In Fig.~\ref{fig:eth_visual}, we list two qualitative examples of our method and baseline models.
In the first row, we see a pedestrian turns right at the end of the ground truth future. 
We notice that Social-Implicit and Trajectron++ cover the ground truth future well, whereas S-GAN and ExpertTraj give us an early turning prediction and concentrate away from the ground truth. 
The second row shows a zigzag walking pedestrian. Baseline models like S-STGCNN, Trajectron++, and ExpertTraj cannot generate good distributions to cover the ground truth trajectory, unlike ours and S-GAN.
Although the prediction of ExpertTraj is close to the ground truth, ExpertTraj is over confident contradicting the ground truth. Qualitative results show that our predicted distribution are better. We also show multi-agent interaction in Fig~\ref{fig:interaction}. ExpertTraj is over-confident missing the ground-truth, S-STGCNN have wide variance with collision, Trajectron++ have ground-truth close to predicted distribution tail, while ours have the right balance. More qualitative results in the appendix.

\section{Conclusions}
We introduced the AMD and AMV metrics that evaluate the distribution generated by a trajectory prediction model. We showed that the BoN ADE/FDE metric gives out an inadequate evaluation of the generated distribution. Based on the objective of AMD/AMV metrics to have a model that generates samples that are close to the ground truth with a tight variance, we introduced the usage of IMLE to train our model, Social-Implicit. We showed that Social-Implicit is a memory efficient model that runs in real-time and relies on several concepts such as Social-Zones, Social-Cell and Social-Loss to enhance the performance. Overall, we invite the motion prediction community to adapt the AMD/AMV to have a better evaluation of their methods.


\clearpage
%
%
\bibliographystyle{splncs04}
\bibliography{egbib}

\appendix

\section{Interactive Demo}

\begin{figure}[ht]
\begin{center}
\includegraphics[width=\linewidth]{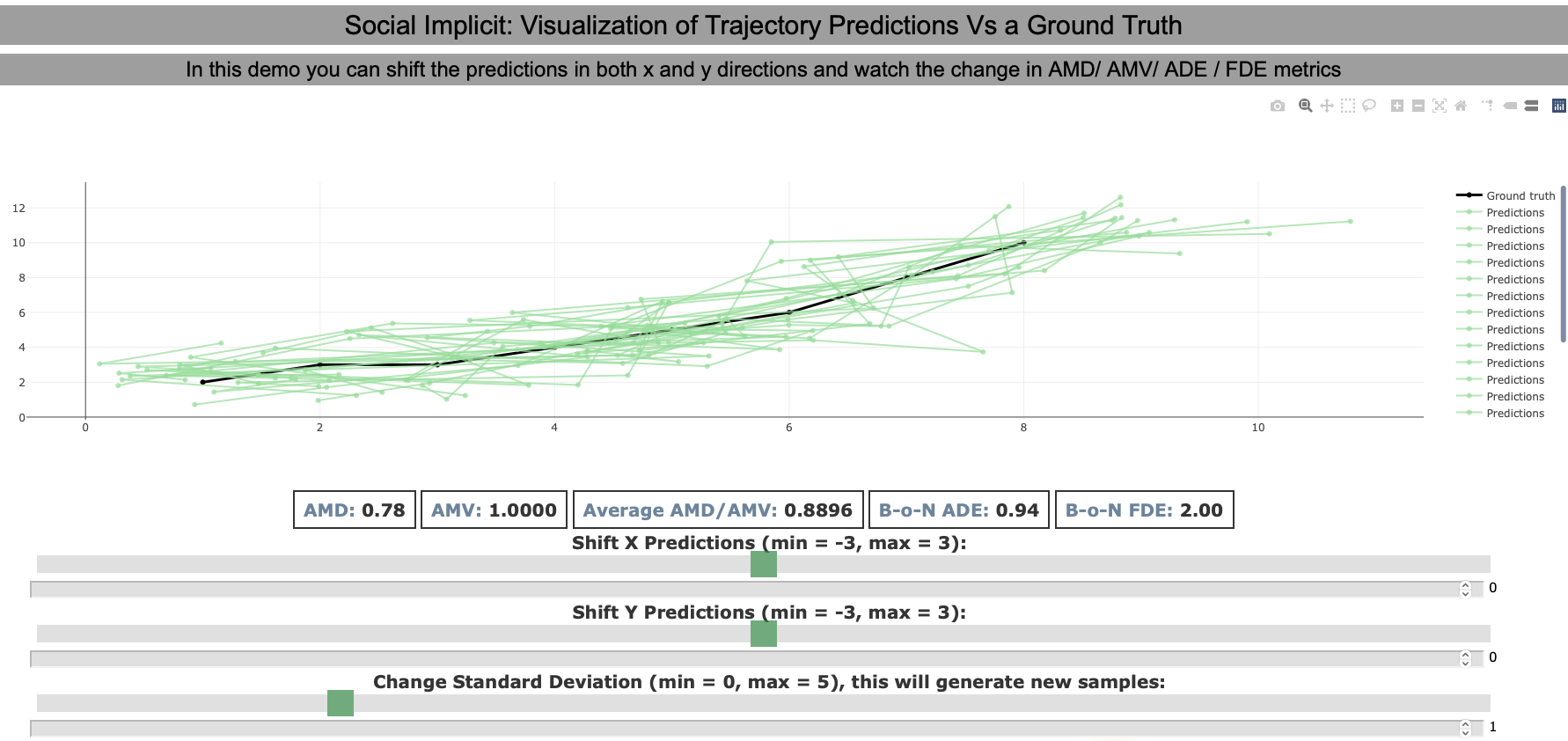}
\caption{Social-Implicit Interactive Demo. This demos shows the changes in the metrics in regards of the generated distribution. }
\label{gr:social_implicit_demo}
\end{center}
\end{figure}
We introduce an interactive demo that shows the change of ADE, FDE, AMD and AMV when the generated distribution changes or shifts. The demo URL: \url{https://www.abduallahmohamed.com/social-implicit-amdamv-adefde-demo}. By using this demo, one can see the direct effect of changing the distribution and how the ADE/FDE metrics are inadequate to evaluate the predicted quality. For example, when the shift is huge in one of the x or y directions, the ADE/FDE will stay constant.

\section{Qualitative Analysis}
\begin{figure*}
\centering
\captionof{figure}{Visualization of the predicted trajectories by several models on the ETH/UCY datasets.}
\label{tab:social_implicit_visual}
\begin{tabular}{ll}
\vcell{1)}     & \vcell{\includegraphics[ width=\linewidth, keepaspectratio]{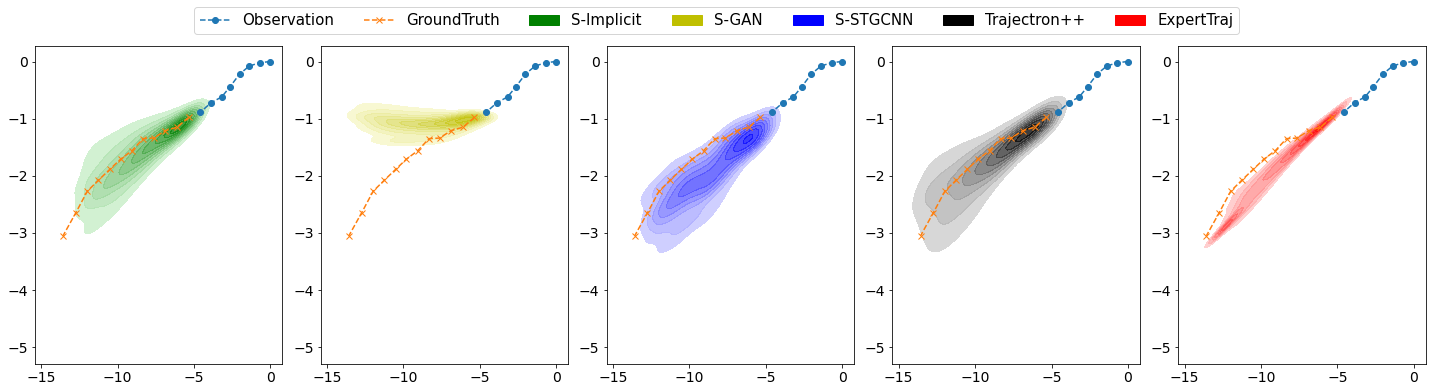}}         \\[-\rowheight]
\printcelltop & \printcellmiddle  \\
\vcell{2)}     & \vcell{\includegraphics[ width=\linewidth, keepaspectratio]{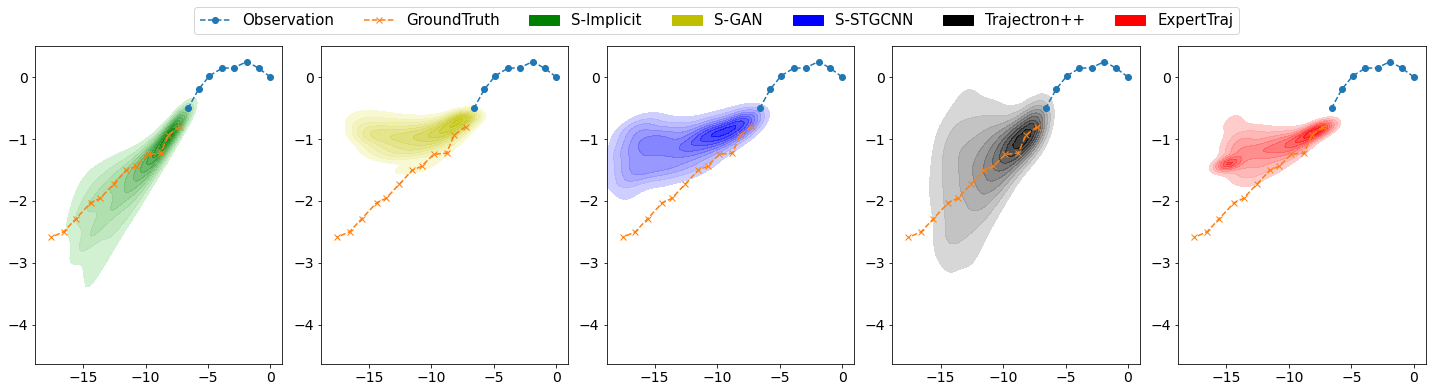}}         \\[-\rowheight]
\printcelltop & \printcellmiddle  \\
\vcell{3)}     & \vcell{\includegraphics[ width=\linewidth, keepaspectratio]{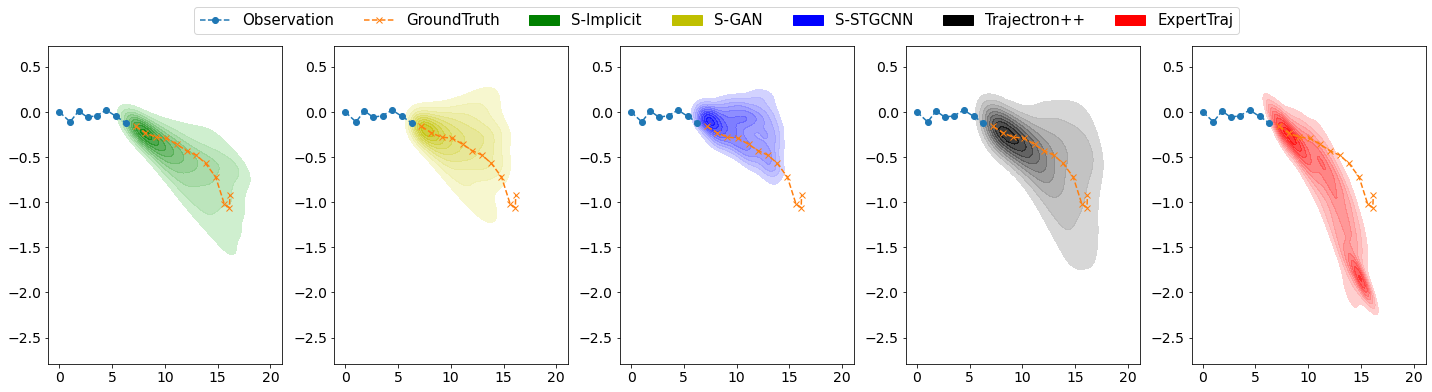}}         \\[-\rowheight]
\printcelltop & \printcellmiddle  \\
\vcell{4)}     & \vcell{\includegraphics[ width=\linewidth, keepaspectratio]{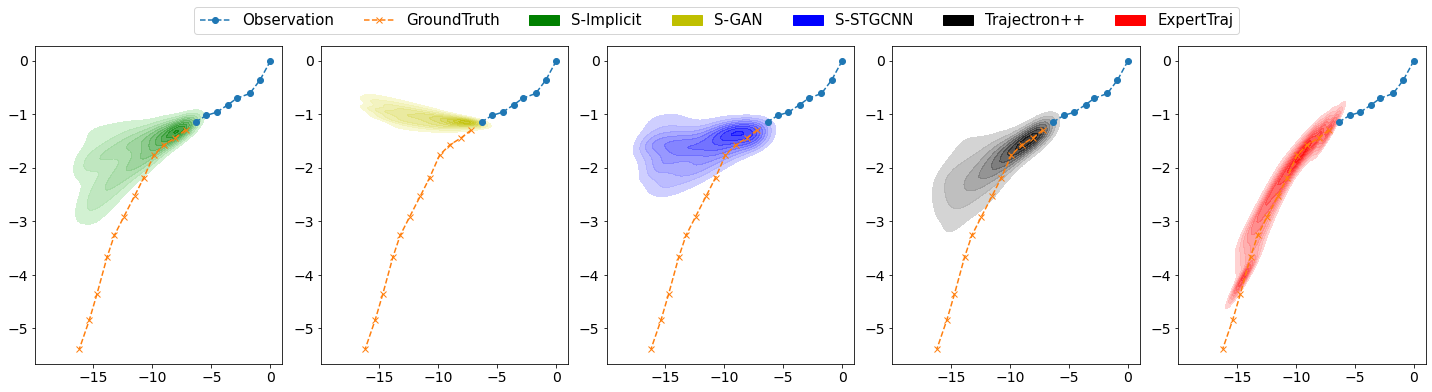}}         \\[-\rowheight]
\printcelltop & \printcellmiddle 
\end{tabular}

\end{figure*}

\begin{figure*}
\centering
\captionof{figure}{Visualization of the predicted trajectories by several models on the ETH/UCY datasets. }
\label{tab:social_implicit_visual_failure}
\begin{tabular}{ll}
\vcell{1)}     & \vcell{\includegraphics[ width=\linewidth, keepaspectratio]{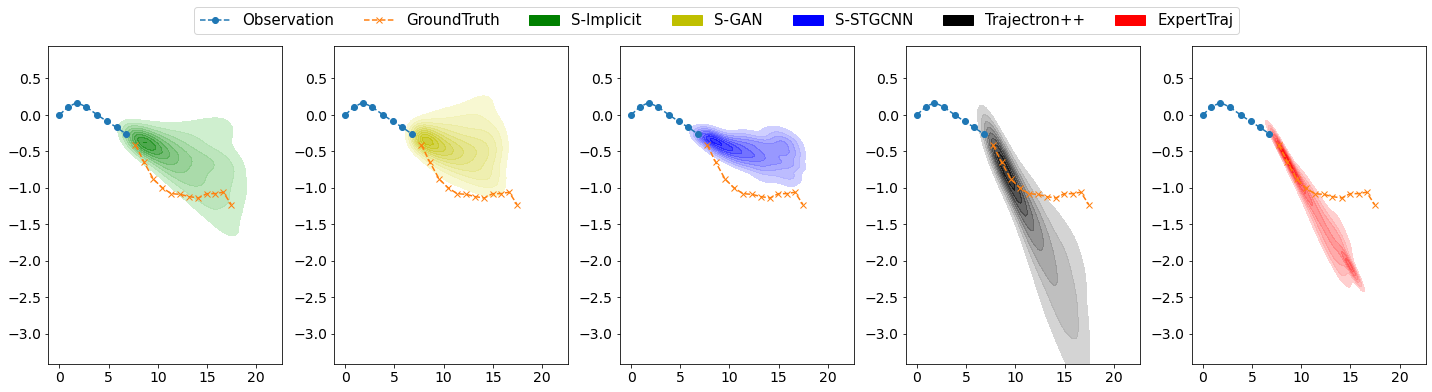}}         \\[-\rowheight]
\printcelltop & \printcellmiddle  \\
\vcell{2)}     & \vcell{\includegraphics[ width=\linewidth, keepaspectratio]{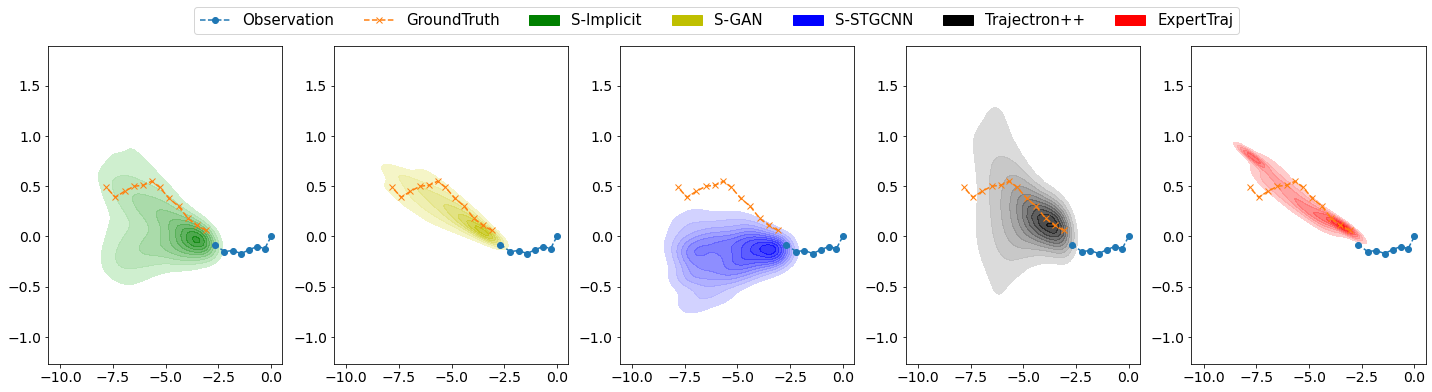}}         \\[-\rowheight]
\printcelltop & \printcellmiddle  \\
\vcell{3)}     & \vcell{\includegraphics[ width=\linewidth, keepaspectratio]{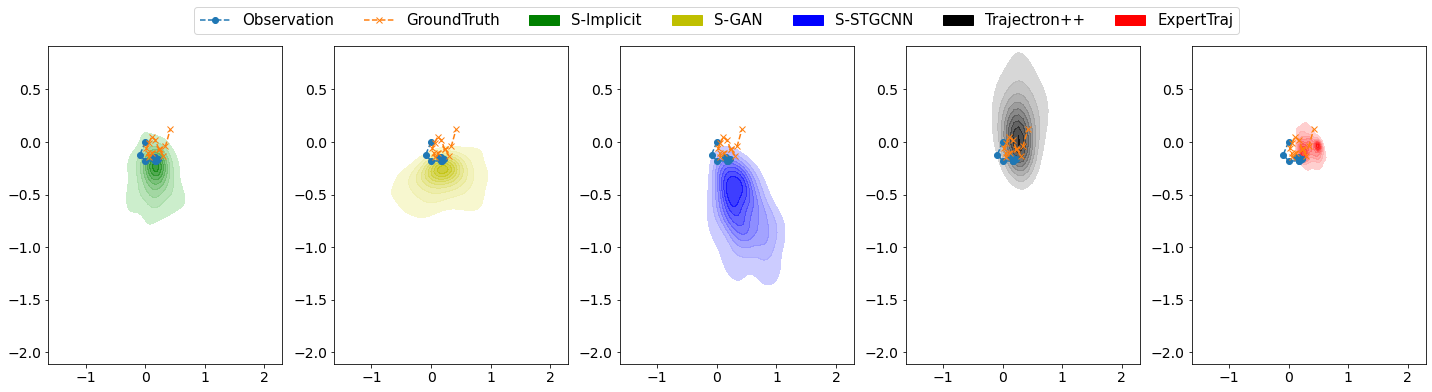}}         \\[-\rowheight]
\printcelltop & \printcellmiddle \\
\vcell{4)}     & \vcell{\includegraphics[ width=\linewidth, keepaspectratio]{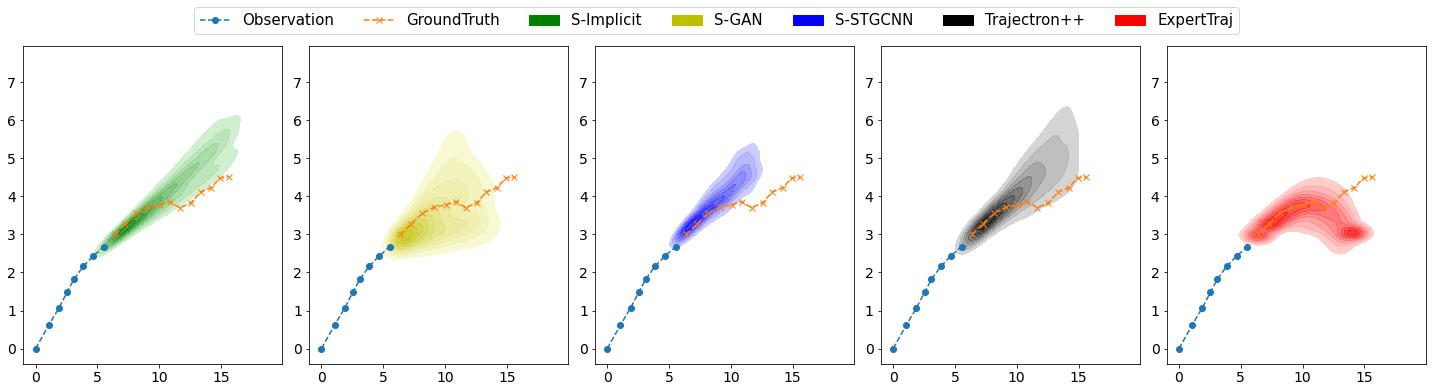}}         \\[-\rowheight]
\printcelltop & \printcellmiddle 
\end{tabular}

\end{figure*}

Figures~\ref{tab:social_implicit_visual} and~\ref{tab:social_implicit_visual_failure} show cases where our model performs well or where it might be under-performing. 
Starting from Figures~\ref{tab:social_implicit_visual}, in the first row and the second row, we see a pedestrian turning left in the past and going straight in the future. S-GAN in the first case and S-GAN, S-STGCNN, ExpertTraj in the second case think confidently that the pedestrian will turn right in the future. But the pedestrian actually goes straightly, which is correctly predicted by our method and Trajectron++. 
In the third case, S-GAN and S-STGCNN give us a too slow prediction and ExpertTraj gives us a too fast and overturning prediction. In contrast, the predicted distribution of our method and Trajectron++ covers the ground truth well.
In the last case, ExpertTraj performs well by placing the predicted concentration on the ground truth. Ours has a wrong prediction following the original trend of the observed motion. 
In the first row and second row of the second Figure~\ref{tab:social_implicit_visual_failure}, the pedestrian has a sudden turn in the middle of the future. Although all the methods fail to predict this turning, the predicted distribution generated by our method covers the ground truth the best.
The third row shows a pedestrian not moving. All the methods give us a close-to-no-movement prediction here. Among them, the movement of ExpertTraj is the smallest. Ours are second-smallest. 
The last row shows a pedestrian going straight but switching the lane in the middle of the future. Our method and Trajectron++ cover the ground truth trajectory well, while S-STGCNN misses the new lane and ExpertTraj generates a no-existing turn.
Overall, though ADE/FDE metrics were stating that Trajectron++ and ExpertTraj are state of art methods, we showed several cases that show the density away from the ground truth. Thus, the ADE/FDE gives an inadequate sense of models' accuracy, unlike the AMD/AMV metrics, which quantifies the whole generated distribution. This correlates with the results in Table[1] and our analysis of the experiments section where our model was performing the best on the ADM/AMV metrics.

\section{Evaluation of Deterministic Models}
\begin{figure}[h]
\centering
\begin{tabular}{@{\hskip1pt}l@{\hskip1pt}l@{\hskip1pt}}
\includegraphics[ width=0.49\columnwidth]{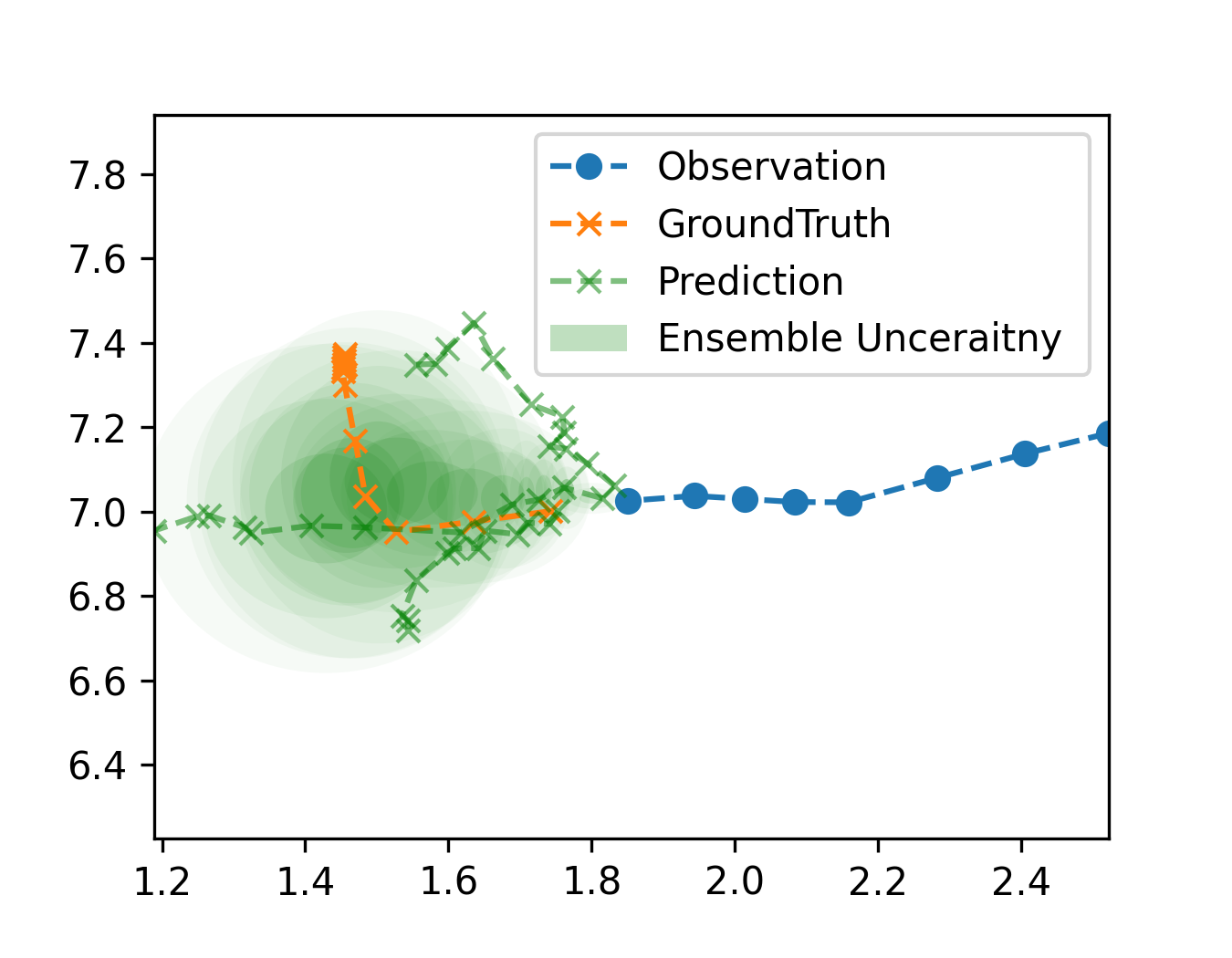} & \includegraphics[ width=0.49\columnwidth]{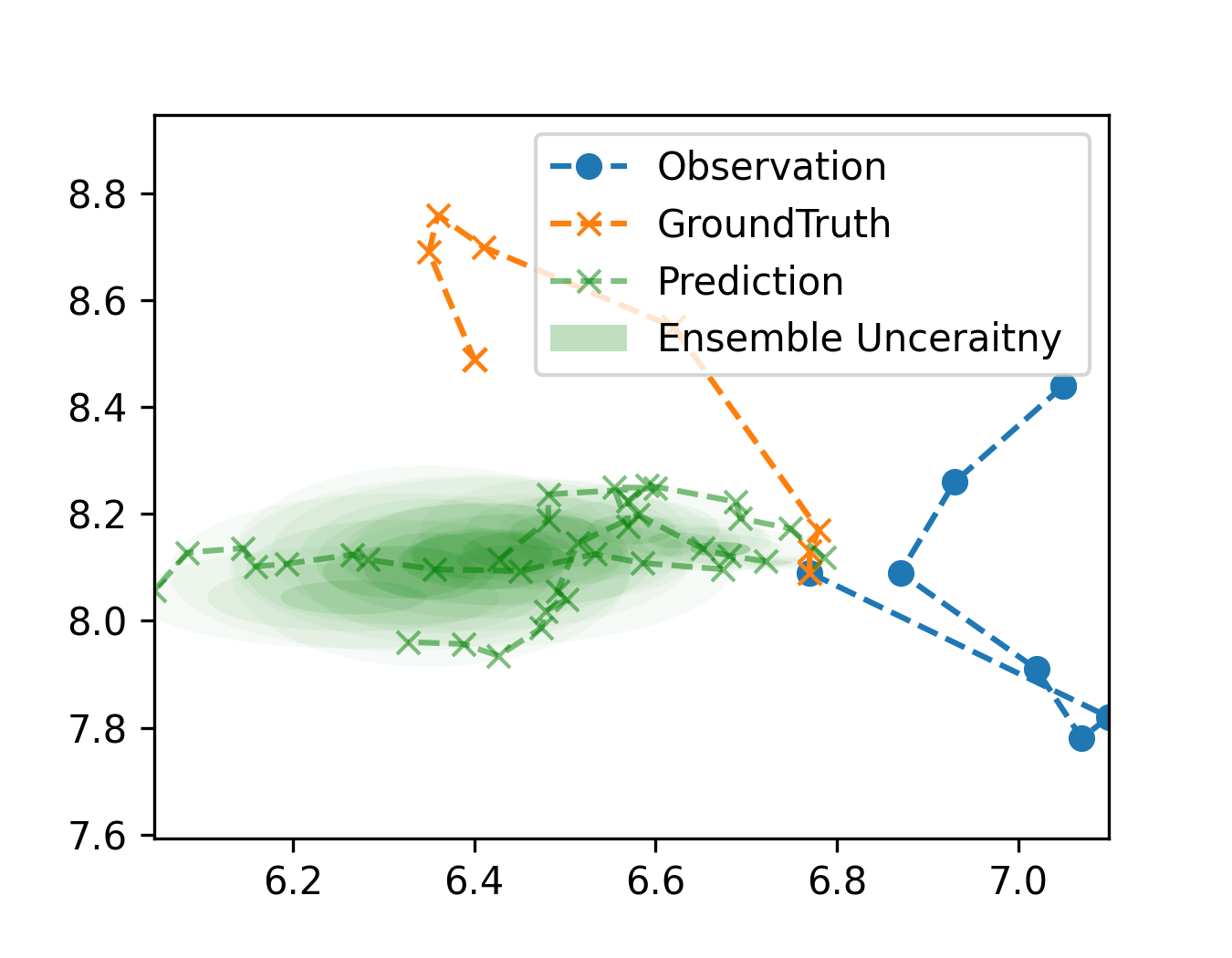}  \\
\includegraphics[ width=0.49\columnwidth]{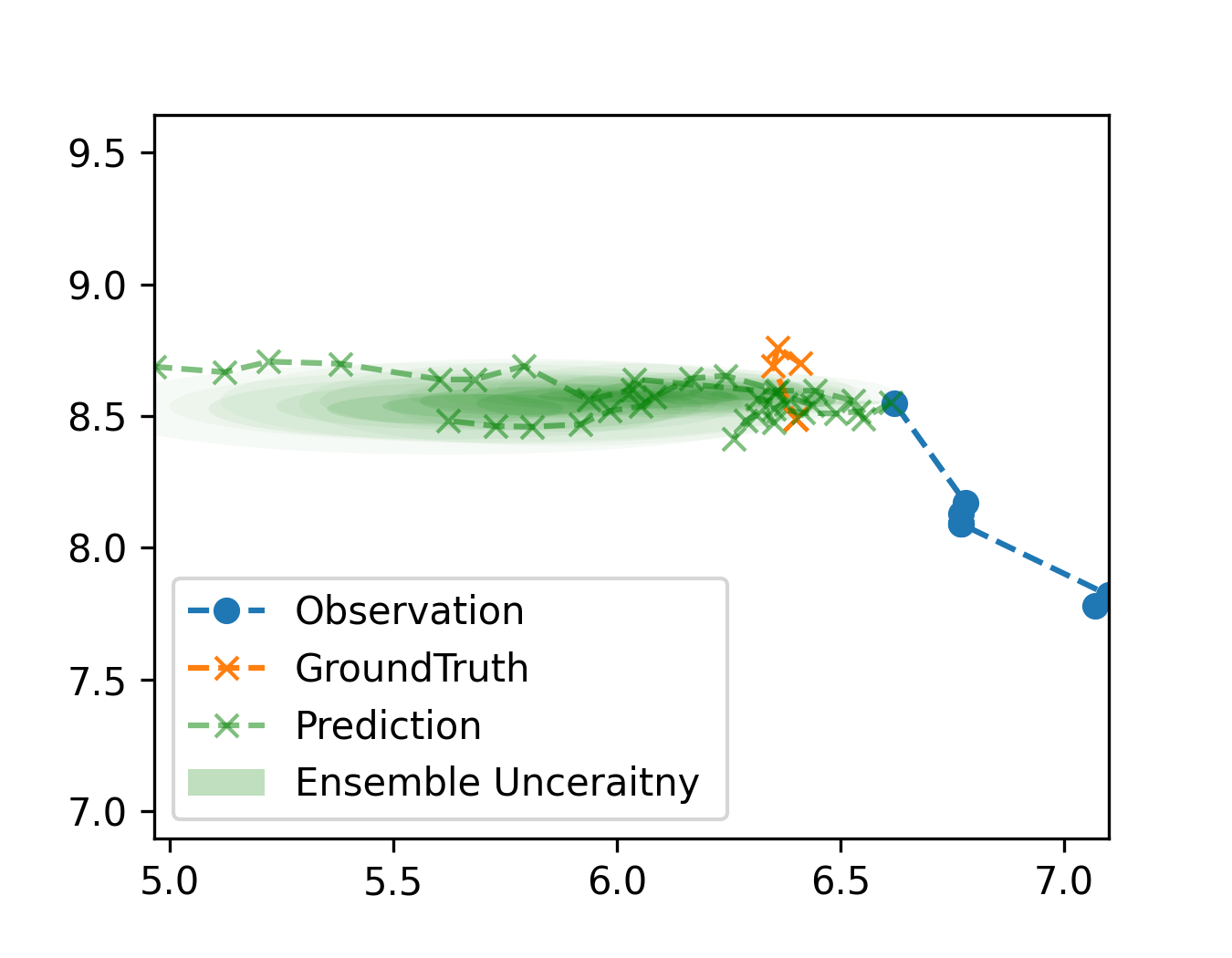} & \includegraphics[ width=0.49\columnwidth]{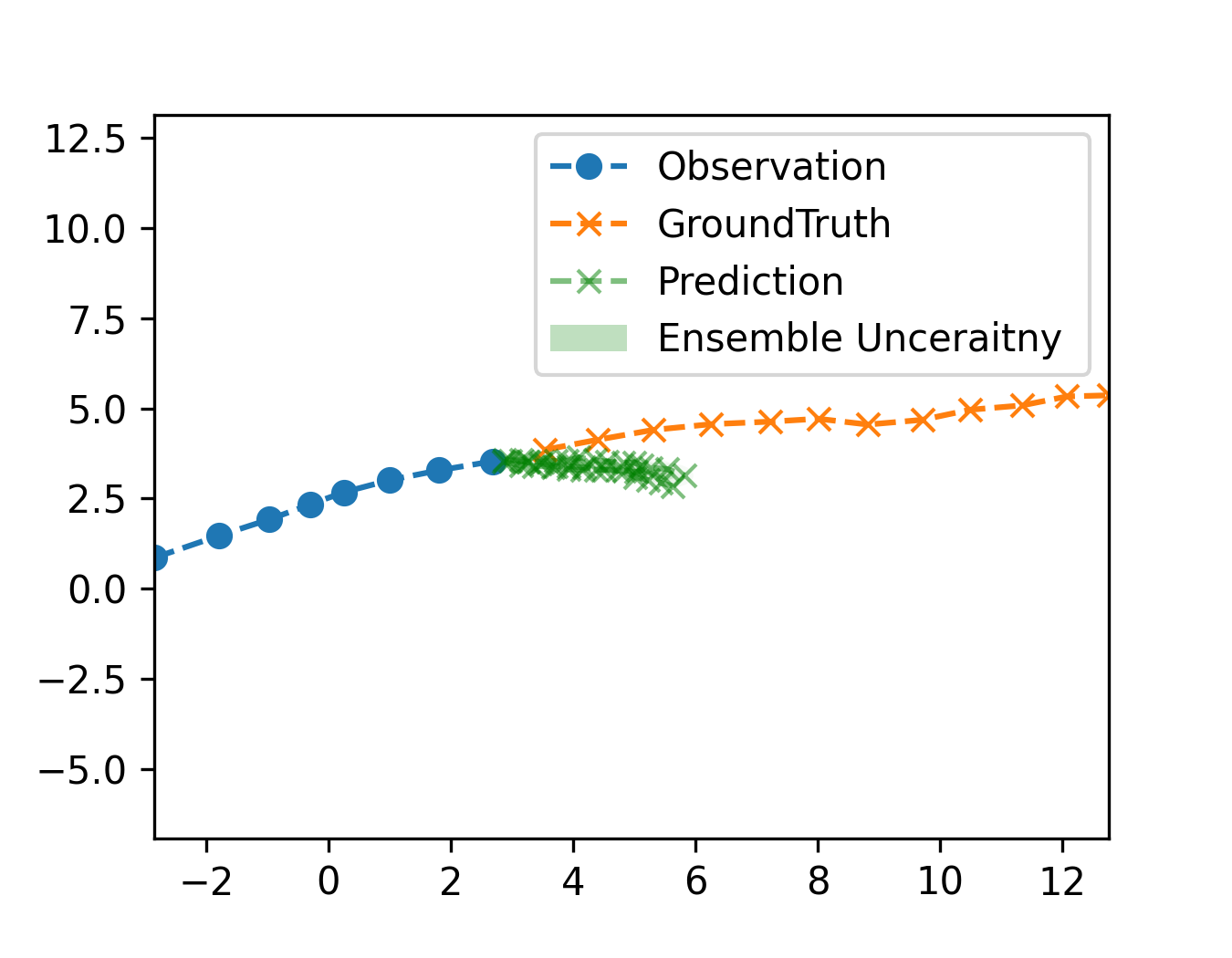}  
\end{tabular}
\captionof{figure}{Social-STGCNN deterministic version predictions. }
\label{tab:social_implicit_visual_det}
\end{figure}

We trained Social-STGCNN~\cite{mohamed2020social} as a deterministic model on the ETH/UCY datasets. Instead of predicting a Gaussian distribution, it predicts the trajectory directory. The training used MSE as a loss function. We wanted to test two assumptions for evaluating a deterministic model. The first one is to train it multiple times and use ensemble to find the mean and variance per predicted trajectory. The other one adds up on the previous one by calculating the mean and variance but fits a GMM and then samples from this GMM. In this experiment, we trained Social-STGCNN 3 times using different random seeds. Table~\ref{tab:social_implicit_determinsitic} shows the results. The AMD and AMV of the first setting was reported. The KDE was not because there is no method to compute it from a mean and variance without sampling, unlike our metric AMD which has this ability by directly plugging in the mean and variance into the Mahalanobis distance equation. For the second setting, AMD, AMV and KDE were reported as we fitted the samples into a GMM fit then we sampled multiple samples. We only used 3 ensembles of Social-STGCNN to simulate a real-life situation, aka it is not feasible to train it 1000 times and create an ensemble out of it. We notice in Table~\ref{tab:social_implicit_determinsitic} that the second settings exhibit a very large AMD and KDE, this is an indicator that the GMM fit did not converge because we only have 3 samples. Usually, we use 1000 samples to guarantee the GMM converges and thus the second settings is not feasible to be used as we need that many samples to fit the GMM model well. We notice in both first and second settings that the AMV values are the same. This was expected, as the AMV metric is an indicator of the spread. For the first setting, the AMD value seems reasonable for a deterministic model, as the work of~\cite{makansi2021exposing} showed that most of motion predictions problem can be solved using a linear Kalman filter. This also supported by the enormous values of the AMV metric as a deterministic model does not have that much of a spread. We connect this with the results in the main paper on the ExpertTraj where the AMV values was on the same order of magnitude as the deterministic model we trained. In other terms, the ExpertTraj indeed behaves as a deterministic model because of the tight spread. We can notice this in some of the visual cases reported in Figures~\ref{tab:social_implicit_visual} and~\ref{tab:social_implicit_visual_failure}. For further analysis, we plot some samples generated from the ensemble of the deterministic model alongside the spread in Figure~\ref{tab:social_implicit_visual_det}. We notice sometimes the spread of the predictions might be close to the ground truth as in the sample in the top left corner. Also, it can be completely off, as in the other samples. So we think that using an ensemble of a few versions of a deterministic model is a good approach to evaluate its performance using the AMD/AMV metric. Also, with the AMV metric as a target to optimize for, one can train the ensemble using methods that help encourage diversity~\cite{liu2019deep}.
\begin{table}
\centering
\scriptsize
\begin{tabular}{|l|llc||lll||ll|} 
\hline
\multirow{2}{*}{Dataset} & \multicolumn{3}{c||}{Ensemble} & \multicolumn{3}{c||}{GMM Fit} & \multicolumn{2}{c|}{General}  \\ 
\cline{2-9}
                         & AMD  & AMV   & \multicolumn{1}{l||}{KDE}    & AMD   & AMV   & KDE                                  & ADE  & FDE                    \\ 
\hline
eth                      & 1.45 & 35.7  & -                            & 27.89 & 35.7  & 12.89                                & 1.71 & 2.97                   \\
hotel                    & 0.36 & 19.6  & -                            & 44.26 & 19.5  & 11.27                                & 1.41 & 2.56                   \\
univ                     & 0.62 & 170.1 & -                            & 24.62 & 169.9 & 13.30                                & 1.17 & 2.13                   \\
zara1                    & 1.18 & 28.0  & -                            & 28.95 & 28.0  & 13.86                                & 1.71 & 3.18                   \\
zara2                    & 1.03 & 96.9  & -                            & 12.54 & 96.8  & 8.36                                 & 1.16 & 2.10                   \\ 
\cmidrule[\heavyrulewidth]{1-7}\cmidrule{8-9}\cmidrule{8-9}
Average                  & 0.93 & 70.06 & -                            & 27.65 & 70.0  & 11.94                                & 1.43 & 2.59                   \\
\hline
\end{tabular}
\caption{Deterministic case experiment. We trained Social-STGCNN~\cite{mohamed2020social} as a deterministic model using different random seeds. The first setting reports the AMD/AMV using the mean and variance of the ensemble. The second setting reports the AMD/AMV/KDE using a GMM fit on the mean and variance of the ensemble. The ADE/FDE are the average through the ensembles. }
\label{tab:social_implicit_determinsitic}
\end{table}

\section{Social-Implicit Implementation Details}
Social-Implicit comprises four zones, as discussed before. Table~\ref{tab:social_implicit_zones} shows more details about the zones. We notice each zone uses a different configuration of the random noise used to generate the samples. The slow zones use noise which has much lower variance than the faster zones. We also show the layer details of the Social-Cell in Table~\ref{tab:social_implicit_config}. Both local and global streams share the same design, except that the local stream uses Conv1D and the global stream uses Conv2D. We initialize the noise, global and local weights to zero. The noise weight is being multiplied by the sample generated from the random distribution and then added to the input tensor. The models were trained for 50 epochs with a learning rate  = 1, then the learning rate drops to 0.1 after 45 epochs. The batch size was set to 128. We used SGD as an optimizer. We also used an augmentation technique for the trajectories similar to~\cite{salzmann2020trajectron++} to fight some imbalance in the datasets. We used random rotation by several degrees, reverse the trajectory, flip the x,y locations, jitter the location by a small value, increase the number of the nodes in the scene by combining it with another scene and changing the speed of the pedestrians. Implementation of the model and augmentation is available in the attached code.

\begin{table}[ht]
\centering
\small
\begin{tabular}{|l|l|l|} 
\hline
Zone & Speed range & Noise~                  \\ 
\hline
1    & 0-0.01 m/s               & $\mathcal{N}(0,0.05^2)$, if eth $\mathcal{N}(0,0.175^2)$ \\
2    & 0.01-0.1 m/s             & $\mathcal{N}(0,1^2)$ if eth $\mathcal{N}(1.5^2)$            \\
3    & 0.1-1.2 m/s              & $\mathcal{N}(0,4^2)$                        \\
4    & 1.2 ms -~                & $\mathcal{N}(0,8^2)$                   \\
\hline
\end{tabular}
\caption{Social-Zones configurations. The speed range determines if an observed trajectory will be withing the zone or not. The random noise exhibits different variances depending on the zone.}
\label{tab:social_implicit_zones}
\end{table}

\begin{table}
\centering
\small
\begin{tabular}{|l|l|l|} 
\hline
Section                             & Layer Name         & Configuration      \\ 
\hline
\multirow{5}{*}{Local Stream      } & Spatial CNN        & Conv1D[$P$,$P$,3,1]    \\
                                    & Spatial Activation & ReLU               \\
                                    & Spatial ResCNN     & Conv1D[$P$,$P$,1,0]    \\
                                    & Temporal CNN       & Conv1D[$T_o$,$T_p$,3,1]  \\
                                    & Temproal ResCNN    & Conv1D[$T_o$,$T_p$,1,0]  \\ 
\hhline{===|}
\multirow{8}{*}{Global Stream}      & Noise Weight            & 1 Parameter        \\
                                    & Spatial CNN        & Conv2D[$P$,$P$,3,1]    \\
                                    & Spatial Activation & ReLU               \\
                                    & Spatial ResCNN     & Conv2D[$P$,$P$,1,0]    \\
                                    & Temporal CNN       & Conv2D[$T_o$,$T_p$,3,1]  \\
                                    & Temproal ResCNN    & Conv2D[$T_o$,$T_p$,1,0]  \\
                                    & Global Weight           & 1 Parameter        \\
                                    & Local Weight            & 1 Parameter        \\
\hline
\end{tabular}
\caption{Social-Cell configuration. A Conv1D or Conv2D with [x,x,x,x] = [input features, output features, kernel size, padding size]. The Res = Residual connection being added to the previous layer output. $P$ is the dimension of the observed location. $T_o$ and $T_p$ is the number of observed and predicted time steps.}
\label{tab:social_implicit_config}
\end{table}

\end{document}